\documentclass{article}

\usepackage{microtype}
\usepackage{graphicx}
\usepackage{subcaption}
\usepackage{booktabs} 

\usepackage{hyperref}

\usepackage[accepted]{icml2026}

\usepackage{amsmath}
\usepackage{amssymb}
\usepackage{mathtools}
\usepackage{amsthm}
\usepackage{tabularx}
\usepackage{array}

\usepackage[capitalize,noabbrev]{cleveref}

\theoremstyle{plain}

\theoremstyle{definition}

\theoremstyle{remark}

\usepackage[textsize=tiny]{todonotes}

\usepackage{wrapfig,lipsum,booktabs}
\usepackage[table]{xcolor}
\newcommand{\gain}[1]{\textcolor{green!50!black}{#1}}   
\usepackage[normalem]{ulem}
\useunder{\uline}{\ul}{}

\newcolumntype{Y}{>{\raggedright\arraybackslash}X}
\newcolumntype{C}{>{\centering\arraybackslash}X}
\renewcommand{\tabularxcolumn}[1]{m{#1}}

\usepackage{graphicx}
\usepackage{subcaption}
\usepackage{pifont}

\icmltitlerunning{Toward Cybersecurity-Expert Small Language Models}

\begin{document}

\twocolumn[
  \icmltitle{Toward Cybersecurity-Expert Small Language Models}




  \begin{icmlauthorlist}
    \icmlauthor{Matan Levi}{yyy}
    \icmlauthor{Daniel Ohayon}{yyy}
    \icmlauthor{Ariel Blobstein}{yyy}
    \icmlauthor{Ravid Sagi}{yyy}
    \icmlauthor{Ian Molloy}{yyy}
    \icmlauthor{Yair Allouche}{yyy}

  \end{icmlauthorlist}

  \icmlaffiliation{yyy}{IBM Research}

  \icmlcorrespondingauthor{Daniel Ohayon}{daniel.ohayon@ibm.com}

  \icmlkeywords{Machine Learning, ICML}

  \vskip 0.3in
]



\printAffiliationsAndNotice{}  

\begin{abstract}
Large language models (LLMs) are transforming everyday applications, yet they lag behind in specialized fields, such as cybersecurity, due to a lack of high-quality, domain-specific models and training datasets. To address this gap, we present \textit{CyberPal~2.0}, a family of cybersecurity-expert small language models (SLMs) ranging from 4B–20B parameters. To train CyberPal~2.0, we generate an enriched chain-of-thought cybersecurity instruction dataset built with our data enrichment and formatting pipeline, \textit{SecKnowledge~2.0}, which integrates expert-in-the-loop steering of reasoning formats alongside LLM-driven multi-step grounding, yielding higher-fidelity, task-grounded reasoning traces for security tasks. Across diverse cybersecurity benchmarks, CyberPal~2.0 consistently outperforms its baselines and matches or surpasses various open and closed-source frontier models, while remaining a fraction of their size. On core threat-investigation tasks, such as correlating vulnerabilities and bug tickets with weaknesses, our best 20B-parameter model \emph{outperforms GPT-4o, o1, o3-mini, and Sec-Gemini~v1}, ranking \emph{first}, while our smallest 4B-parameter model ranks \emph{second}. On core cyber threat intelligence knowledge tasks, our models outperform almost all tested frontier models, ranking \emph{second only to Sec-Gemini~v1}. To foster reproducibility and practical adoption, we will release our models as open source.
\end{abstract}

\section{Introduction}
\label{intro}
Threat management and security operations are a natural fit for language models, because they require reasoning over diverse evidence and security-specific context. One of the most promising directions for practical impact is threat management and security operations \citep{motlagh2024large, yao2024survey}. This setting demands deep, cross-domain understanding spanning network, cloud, and application security, along with the ability to investigate and reason over threat reports \citep{zhang2025llms, lin2025ircopilot}.
In this paper, we focus on this direction and propose a single defensive model that supports the full security-operations loop. Our goal is a domain-specialized backbone that delivers core capabilities for detection, investigation, response, threat hunting, and data classification, while remaining straightforward to integrate and deploy within enterprise pipelines.

However, adopting frontier models for security is a challenging task.
Commercial offerings typically enforce strict safety guardrails, which limit their practical utility in real-world security workflows \citep{weerawardhena2025llama}. Additionally, full integration with organizational data sources is further constrained by compliance requirements, as security data often contains highly sensitive and private information \citep{zhang2025llms}. 
For these reasons, many enterprises require on-premises solutions to meet privacy, compliance, and data residency obligations, making it impractical to send sensitive telemetry to external frontier services. 
Cost is another factor, where models must handle massive volumes of security data efficiently. 
These constraints make security another domain where domain-specific Small Language Models (SLMs) are preferable to general-purpose frontier models \citep{belcak2025slm}. 

In addition to practical deployability, such a model must support core capabilities in the cybersecurity domain. 
It requires deep technical grounding across multiple domains: operating systems, computer networks, cloud platforms, identity and access management, and enterprise security controls \citep{li2021comprehensive, aslan2023comprehensive}.
Most importantly, the model needs to incorporate comprehensive understanding of threats that includes attacker tactics, techniques, and procedures (TTPs), as well as software vulnerabilities, weaknesses, adversary tooling, and probable attack paths. It must also align with security operation workflows, covering hypothesis-driven threat hunting, investigation, and severity assessment.
Finally, such a model must connect the dots, reason effectively over security evidences, and deliver defensible conclusions that are accurate and reliable.

\begin{figure}[t]
  \centering
  \begin{minipage}[t]{0.49\textwidth}
    \centering
    \includegraphics[width=\linewidth]{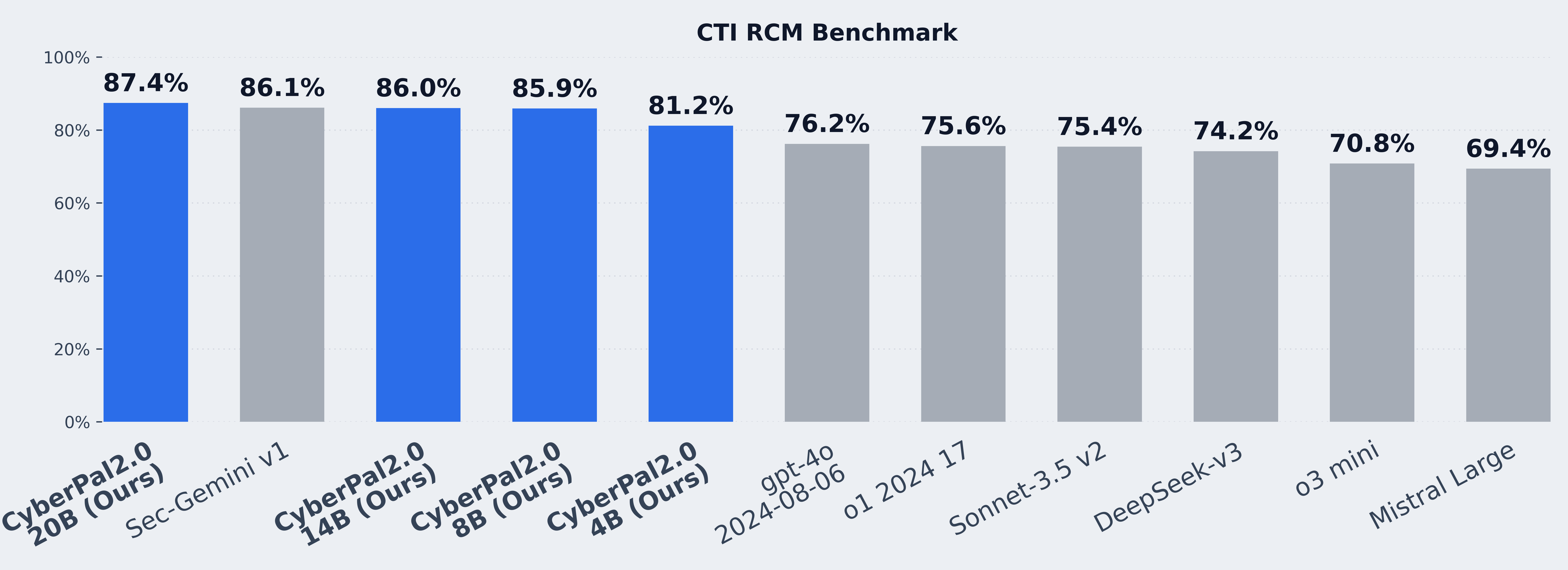}
    {\small (a) Evaluation results on weaknesses Root Cause Mapping tasks.}
  \end{minipage}\hfill
  \begin{minipage}[t]{0.49\textwidth}
    \centering
    \includegraphics[width=\linewidth]{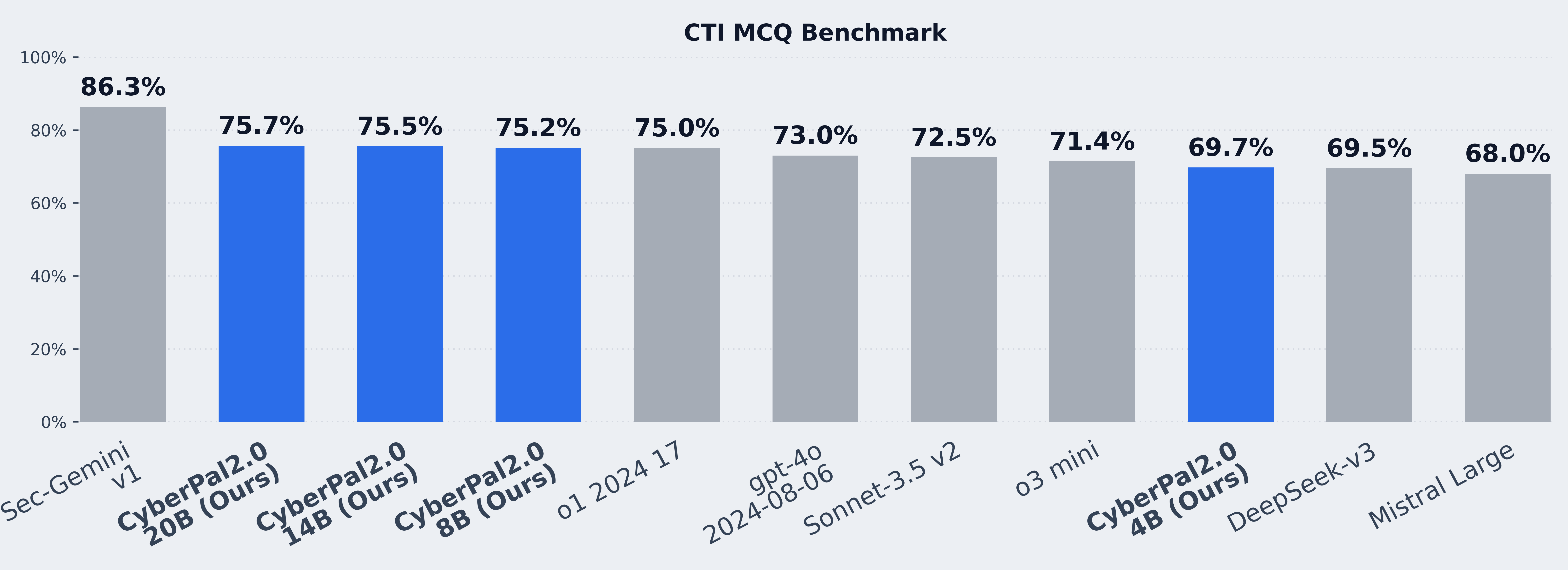}
    {\small (b) Evaluation results on core cyber threat intelligence tasks.}
  \end{minipage}
  \caption{Comparing our models (blue) against frontier models such as Sec-Gemini~v1, o1, and o3-mini (gray) on key benchmarks. }
  \label{fig:sota}
\end{figure}

In earlier work, \citeauthor{levi2025cyberpal} took an initial step in this direction. They introduced SecKnowledge, a domain-knowledge-driven cybersecurity instruction dataset built through a multi-phase generation process anchored in expert curation, along with SecKnowledge-Eval, a comprehensive evaluation suite covering a wide range of cybersecurity tasks. Models fine-tuned on SecKnowledge demonstrated significant improvements over baseline methods, highlighting the effectiveness of expert-guided instruction design and domain-specific evaluation in advancing cybersecurity LLMs.

This work takes a substantial step toward a practical security model for threat management and security operations:
\begin{itemize}
\item \textbf{SecKnowledge~2.0.} We propose a dataset enrichment pipeline that incorporates domain expertise via expert-in-the-loop  schema-driven formatting, and applies multi-source, multi-step grounding to improve reasoning traces for security tasks and overall data quality.

\item \textbf{Suite of cybersecurity-expert SLMs.} We train a suite of cybersecurity-focused SLMs, ranging from 4B to 20B parameters, that reason over complex threats and map domain knowledge to setting-specific analyses and recommendations.

\item \textbf{Frontier cybersecurity performance.} Across rigorous cybersecurity benchmarks, our SLMs consistently outperform their baselines and state-of-the-art open-source models, yielding 7--14\% average gains; on core cybersecurity threat intelligence (CTI) benchmarks, our models match or surpass frontier closed models (e.g., Sec-Gemini~v1, OpenAI's o1), all while retaining the cost efficiency, openness, and on-premises deployability required by enterprises.
\end{itemize}

To support reproducibility and further research, we release the SecKnowledge 2.0 pipeline\footnote{\href{https://github.com/d4nieldev/fms-dgt/tree/secknowledge2/fms_dgt/public/databuilders/secknowledge2}{SecKnowledge 2.0 code}}, CyberPal 2.0 20B model\footnote{\href{https://huggingface.co/cyber-pal-security/CyberPal2.0-20B}{CyberPal 2.0 20B Model (HuggingFace)}}, and our evaluation benchmarks\footnote{\href{https://huggingface.co/datasets/cyber-pal-security/SecKnowledge-Eval}{SecKnowledge-Eval benchmarks suite (HuggingFace)}}.

\paragraph{Conflict of Interest Disclosure.} 

All authors are employed by IBM. The CyberPal 2.0 models introduced and evaluated in this paper were developed at IBM.


\section{Related Work}
\label{related_work}
Recent work positions LLMs as security tools across Cyber Threat Intelligence (CTI), malware analysis, and incident response, among other security tasks.
Recent systematic reviews synthesize both the landscape and open gaps in evaluation and datasets \citep{zhang2025llms, xu2024large}. 
In this work, we focus primarily on  using LLMs as security tools and evaluate their 
performance in applied security settings. 

\citeauthor{yu2025primus} curates a multi-source cybersecurity corpus for pre-training (web content, blogs, books, Wikipedia, and MITRE-linked resources), filters a general crawl for security-related text, and augments it with LLM-style rewrites; it then performs instruction fine-tuning on real-life cybersecurity-oriented tasks with LLM-generated references and distills reasoning on CTI-Bench using a general-purpose model with chain-of-thought. Despite the breadth of the pre-training data, their fine-tuning dataset is limited in size and is derived primarily via distillation.
Following this work, \citeauthor{weerawardhena2025llama} created an instruction-tuned, security-specialized chat model built on the Foundation-Sec-8B base (a Llama-3.1-8B continued-pretrained on a curated cybersecurity corpus), which is competitive with both open and closed models like Gemma \citeauthor{team2025gemma} and GPT-4o-mini \citep{hurst2024gpt}.
The work minimizes security-specific content during post-training, relying on continued pre-training for domain knowledge; their post-training data emphasize diversity and instruction-following rather than security knowledge injection.
Taken together, these studies emphasize security-focused pretraining, leaving underexplored the role of expert-driven, document-grounded supervised fine-tuning, which is crucial not only for reliability but also for enabling practical problem-solving and actionable guidance for cybersecurity workflows \cite{zhang2025llms}.

Few practitioner-built checkpoints appear on community hubs (e.g., Hugging Face) without an accompanying paper or technical report \citep{deephat_v1, lily_cyber_7b_v0_2}.
These releases often omit essential details (e.g., training data sources), making rigorous comparison and reproducibility challenging. Closed vendor security models sometimes likewise report only headline scores. Google’s Sec-Gemini v1 \citep{bursztein2025secgemini} combines Gemini’s reasoning with security knowledge and tools by tying in Google Threat Intelligence (Mandiant/GTI) and OSV\footnote{https://www.mandiant.com/, https://osv.dev/}. They report strong results on CTI  core knowledge and root-cause mapping tasks, though access remains limited.

We build upon SecKnowledge introduced by \citeauthor{levi2025cyberpal}, which takes a data-first route with an expert instruction set and an evaluation suite, and reported sizable improvements in threat-hunting Q\&A and investigation assistance.



\section{SecKnowledge 2.0: Data Reformatting and Enrichment Pipeline}
\label{data_method}

In this section, we introduce \textit{SecKnowledge~2.0} - an enhanced version of \textit{SecKnowledge}, a comprehensive cybersecurity instruction dataset originally introduced by \citeauthor{levi2025cyberpal}, which generates synthetic data from curated cybersecurity seed sets.
\textit{SecKnowledge~2.0} extends \textit{SecKnowledge} via a reformatting and enrichment pipeline, shown to improve downstream task performance  \citep{fan2024reformatted, nguyen2025recycling, phi4}.


This section is organized as follows: Section~\ref{section:SecKnowledge} introduces \textit{SecKnowledge}, which serves as our starting point dataset. Section~\ref{section:realign-original} describes standard data reformatting and enrichment approaches. Section~\ref{pipeline_ext} then presents our improvements on top of the standard approaches described in \ref{section:realign-original}, which combine LLMs with expert-in-the-loop feedback to define reasoning structures and employs LLM-automated query generation to retrieve external evidence for enriched, reliable responses. We use gpt-oss-120b with \textit{Medium} reasoning effort as the backbone LLM. The result is \textit{SecKnowledge 2.0}, a dataset whose responses are structured, interpretable, and supported by evidence.


\begin{figure*}[h]
  \centering
  \includegraphics[width=\textwidth]{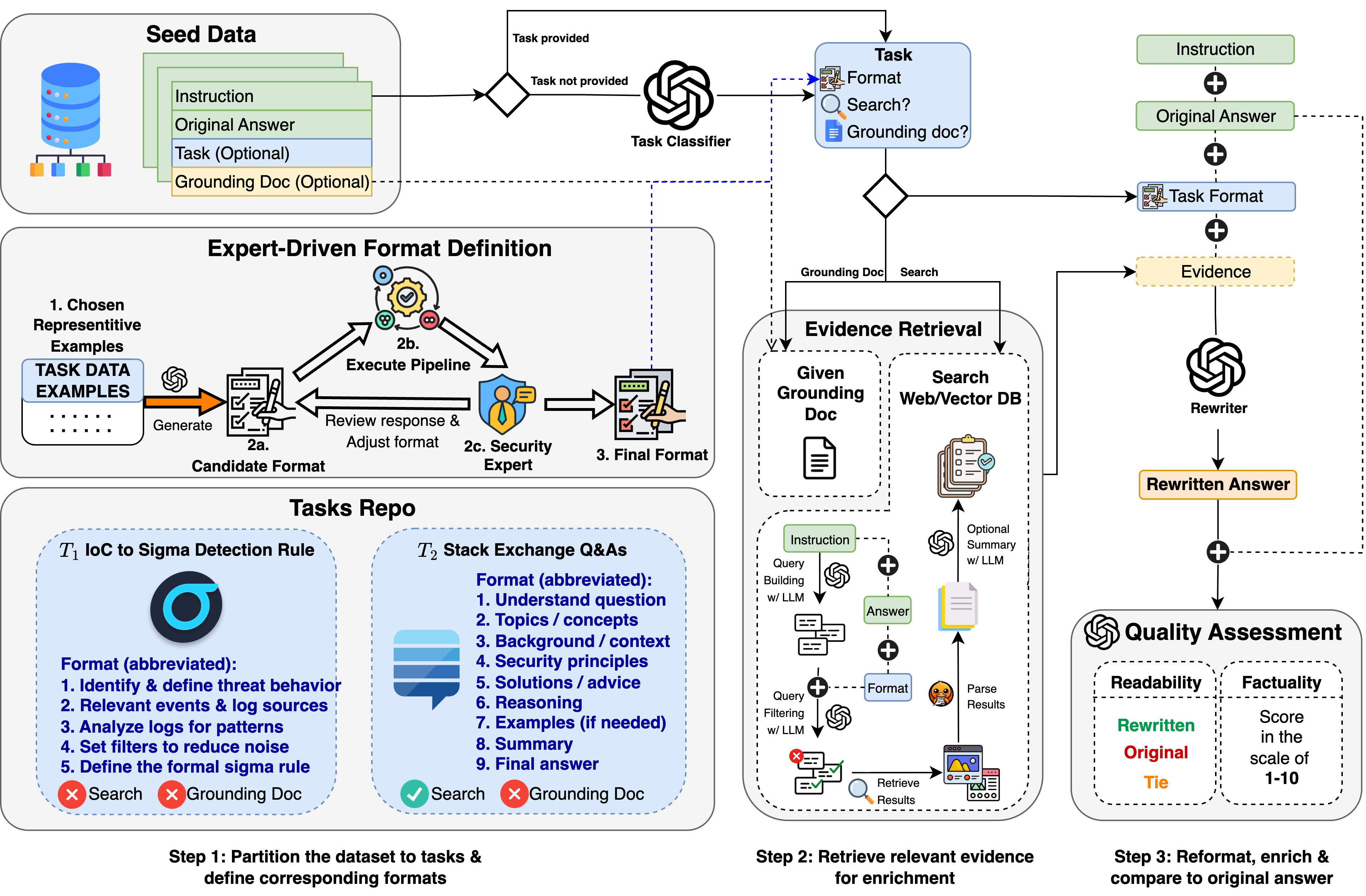}
  \caption{3-step reformatting and enrichment pipeline overview - experts define task formats for the seed dataset, relevant evidence is retrieved from documents or web sources, answers are reformatted and enriched into structured, knowledge-grounded outputs with LLMaaJudge verification (see \ref{pipeline_quality_assessment}).}
  \label{figures:data-pipeline}
\end{figure*}

\subsection{SecKnowledge: A Diverse Set of Cybersecurity Instructions Set}
\label{section:SecKnowledge}

\textbf{SecKnowledge} is a domain-knowledge–driven instruction dataset for cybersecurity, constructed in two stages that combine expert curation with structured automation.
In the first stage, schema-based parsers were designed for foundational security corpora from public security data sources.
In the second stage, SecKnowledge was then extended by generating high-quality seed instructions that capture both per data-source concepts and cross data-source relationships using a novel synthetic data generation method. 
For example, paths in BRON (a graph that interconnects security entities introduced in \cite{hemberg2020linking}) are transformed into chain-of-thought (CoT) \cite{wei2022chain} rationales. 
Sigma rules are converted into step-by-step ``how to detect'' explanations, and SIEM rules are mapped to ATT\&CK TTPs with grounded rationale.
First stage yields $\sim$153k instructions across sources, providing a structurally diverse and practically grounded seed set. In the second stage, \textit{SecKnowledge} increased both diversity and difficulty through dynamic content-grounded synthetic generation, yielding a ~403k-example cyber-security corpus. 

While \textit{SecKnowledge} provides broad coverage and high-quality supervision, instruction families are intentionally template-based, which can yield limited rationales and shorter reasoning chains. Building on this foundation, our work enriches those items with explicit, step-by-step trajectories and stronger grounding
by composing and adapting data reformatting and enrichment methods to the security domain, culminating in \textbf{SecKnowledge~2.0}.

\subsection{Baseline: Data Reformatting and Enrichment Pipeline}
\label{section:realign-original}
We build on prior work showing that reformatting existing data sources, conditioned on chain-of-thought (CoT) reasoning, improves performance on downstream tasks \citep{fan2024reformatted, nguyen2025recycling, phi4} and training token efficiency \citep{team2025kimi}.
These pipelines often incorporate a stage that reformats raw answers into CoT reasoning traces, encouraging systematic reasoning. 
Within this line of work, \citeauthor{fan2024reformatted} introduces Reformatted Alignment (ReAlign), a format-driven pipeline that upgrades instruction datasets through three stages: (i) humans define CoT formats; (ii) enrichment adds auxiliary information; and (iii) reformatting imposes an explicit CoT structure.

Despite \textit{SecKnowledge}’s breadth, responses tend to be concise with short rationales. 
A pipeline such as ReAlign can expand these compact answers into explicit, step-by-step trajectories while grounding them in retrieved documents and authoritative sources, making it a natural baseline and a strong foundation for \textit{SecKnowledge~2.0}. At the same time, instruction generation in complex domains is prone to hallucinations  \citep{jiang2023active} and divergence from expert intent \citep{levi2025cyberpal,ramjee2025cataractbot,eachempati2025integrating}. We therefore believe a more domain-appropriate pipeline for cybersecurity should be adopted.



\subsection{Pipeline Extensions}
\label{pipeline_ext}
In the next section, we move beyond vanilla reformatting by introducing an expert-in-the-loop workflow that semi-automatically derives domain-specific formats for each task in the dataset.
These formats specify the exact reasoning steps needed to reach the final answer. We further enhance them with document grounding and targeted web search, ensuring that each step is anchored in evidence and minimizing hallucinations during reformatting.

\subsubsection{Expert-in-the-Loop System for Automating Domain-Specific Formats}
\label{section:format-generation}

For reformatting and enriching a given dataset \(D\), the first step is to partition it into distinct tasks \(\{ T_1, ..., T_N \}\), each task representing a coherent sub-domain or capability. Formally:
$S = \{T_1,...,T_N | \bigcup_{k=1}^{k=N}T_k=D, T_i \cap T_j = \varnothing \}.$
Since different problem types demand different ways of structuring outputs, each task $\mathbf{T_i}$ must be paired with a corresponding format $\mathbf{F_i}$ (refer to Figure~\ref{figures:format-example} for an example format) that defines the task more precisely by specifying the steps needed to be taken to provide a detailed and logically coherent answer. Manually constructing such tailored formats, however, can be highly time-consuming, particularly in specialized domains such as cybersecurity, where expert knowledge is required, yet remains both scarce and costly.


To efficiently scale format definition across a large and hierarchical label space - such as SecKnowledge, which contains 105 unique tasks - we developed an expert-in-the-loop system capable of semi-automatically generating and evaluating format templates. 
The system employs a LLM that, given a concise task description together with an optional set of illustrative instruction–response examples from any task, generates a corresponding candidate output format. 
Within the same framework, experts can immediately evaluate this format by executing the full pipeline on representative inputs, obtaining rewritten responses along with auxiliary feedback such as search results and LLM-as-a-judge scores for readability and factuality. Based on this feedback, experts can directly edit the format and rerun the pipeline, enabling a tight feedback loop that supports iterative refinement while substantially reducing the manual burden of format specification and enhancing the efficiency, accuracy, and scalability of the pipeline. For implementation details, refer to Appendix \ref{format_generation_framework}.



\subsubsection{LLM-Guided Search and Document Grounding Pipelines}
\label{section:llm-search}

The vast majority of tasks in SecKnowledge can greatly benefit from enrichment through evidence retrieval. Such grounding is necessary to reduce the risk of LLM hallucinations when rewriting responses and to ensure that outputs remain accurate and reliable. Evidence can be provided in two primary ways: (1) by attaching a grounding document directly to the instruction–response pair, such as an advanced persistent threat (APT) report describing a specific attack, or (2) by searching for relevant documents on demand. The first method is largely straightforward, as it simply links an instruction–response pair to the document from which it was derived. The second method, however, requires more substantive mechanisms, such as searching a pre-indexed corpus (e.g., via a vector database) or the world wide web. Accordingly, the discussion that follows concentrates on the latter, given its broader applicability and scalability.
To obtain high-quality search results, we design the mechanism as a structured multi-step process:

\begin{enumerate}
    \item{\textbf{Query building.} Given only the instruction, the LLM is prompted to generate \(K\) candidate search queries. This stage can be viewed as a brainstorming step to provide diverse queries.}
    \item{\textbf{Query filtering.} A second LLM, conditioned on the instruction, the original answer, the task format, and the candidate queries, selects only those queries expected to provide new or useful information that can fill gaps.}
    \item{\textbf{Results retrieval.} The filtered queries are then executed against either a vector database or the web, with the top \(R_{max}\) results retrieved for each query (yielding \(\le K \times R_{max}\) results).}
    \item{\textbf{Results parsing.} For each query, the top \(R\) results that could be parsed to text are retained, while un-processable ones (e.g., websites that block automated access) are discarded.}
    \item{\textbf{Optional summarization.} When retrieved documents are large (e.g., web pages), they may be summarized in a manner conditioned on the task format, ensuring that the retained content aligns with the information required to populate the format steps.}
\end{enumerate}

In our experiments, we used \(K = 2\), \(R_{max} = 8\), and \(R = 2\), without applying summarization. This configuration is motivated by two considerations: (1) automatic summarization often omits critical details required by the task format, and (2) full documents, when not summarized, can accumulate into a large number of tokens that risk exceeding the context window allocated for the LLM. Moreover, even when the context window is not exceeded, long inputs can lead the model to disproportionately attend to the beginning and end of the prompt rather than its middle content \citep{liu2023lost}, effectively nullifying the benefit of providing additional context. To balance these constraints, we restricted retrieval to at most four results (\(K \times R = 4\)), thereby prioritizing fewer, higher-quality search results over larger volumes of noisier content.

\begin{figure*}[h]
  \centering
  \includegraphics[width=0.91\textwidth]{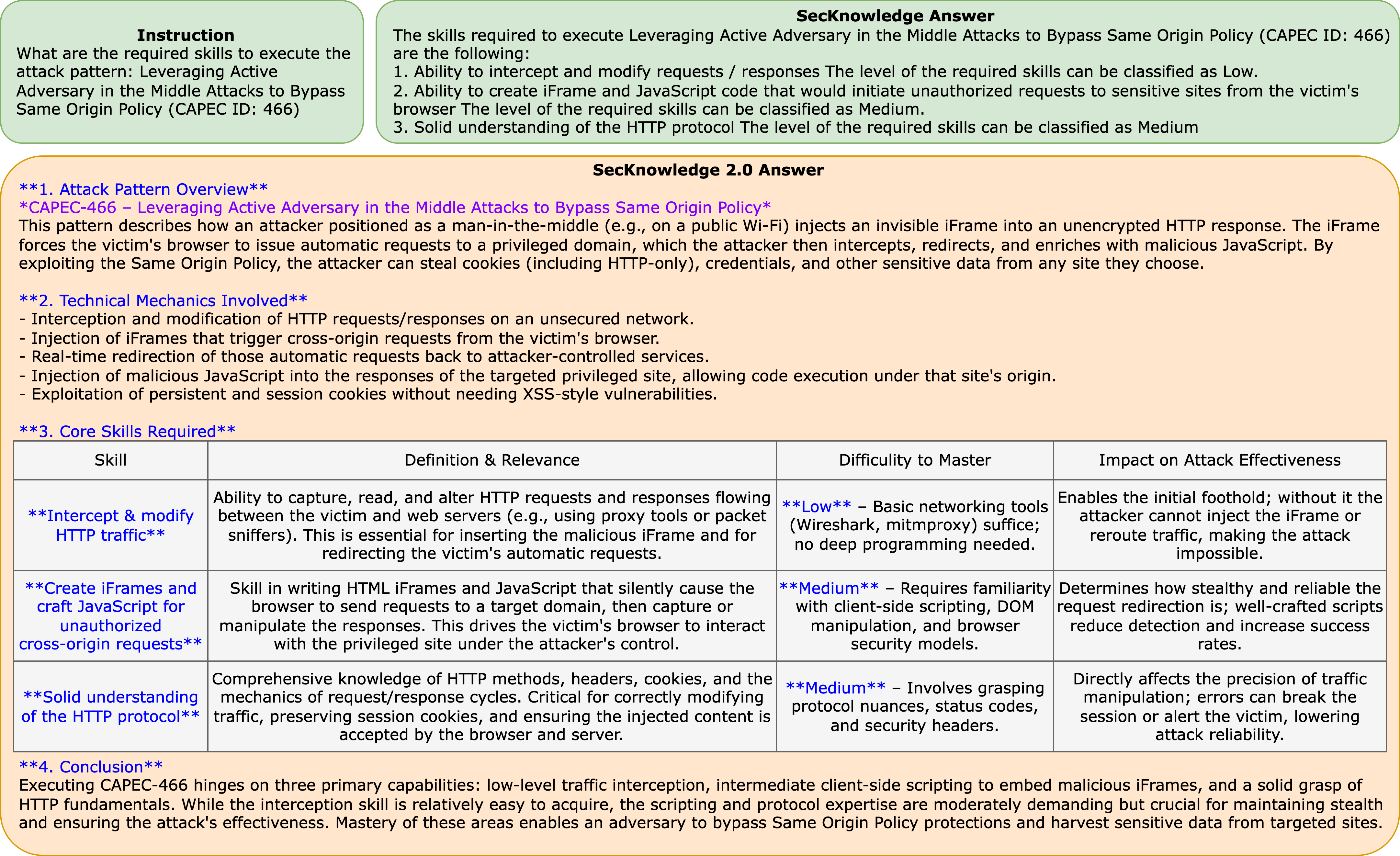}
  \caption{Q\&A example from SecKnowledge (green), and our improved answer (orange).}
  \label{figures:realign-example}
\end{figure*}

\section{Training and Evaluation Process}
\label{cyberpalv2}

We detail the training and evaluation pipelines. For the evaluation, we report results on widely used open-source security benchmarks. When comparing to state-of-the-art frontier models (such as Sec-Gemini v1), we follow the evaluation protocol of Sec-Gemini v1 and report results on the well-known CTIBench-RCM and CTIBench-MCQ \citep{ctibench}. Additionally, for both the baseline and our fine-tuned models, we evaluate on further benchmarks \citep{sec_eval, tihanyi2024cybermetric, levi2025cyberpal}, as elaborated in Section~\ref{eval_benchmarks}.


\subsection{Training Recipe}
\label{train_recipe}
To train our models, we use our generated \textit{SecKnowledge~2.0} dataset. We employ Qwen3-4B-base, Qwen3-8B-base, and Qwen3-14B-base, alongside gpt-oss-20b as our starting point. Training is performed with a learning rate of $4 \times 10^{-5}$ and a linear warm-up ratio of 0.15. The context length is set to 8192, and the batch size is 3072. We train our models for two epochs.

To train our models with adaptive reasoning capabilities, we incorporate adaptable reasoning depth: long-form chain-of-thought examples from SecKnowledge~2.0 are augmented with ``step-by-step'' requests, while shorter instructions from the original SecKnowledge dataset are paired with concise, fast-response requests. This design, similar to the notion of reasoning effort in gpt-oss, balances reasoning-intensive and lightweight tasks. For the shorter, fast-response requests, we sample approximately 25\% of the original instructions and responses from the original SecKnowledge dataset, focusing primarily on short, high-quality examples selected using LLMaaJ. The procedure of mixing a portion of the high-quality original responses with their enhanced counterparts not only teaches the models to perform adaptive reasoning, but also improves token utility by amplifying the volume of high-quality tokens while reducing overfitting, as observed by \citet{team2025kimi}.

Additionally, we observed a phenomenon also reported in recent studies \citep{huerta2024instruction, shi2024instruction, chatterjee2025effect}: it is often preferable to retain at least partial loss on the prompt rather than masking it out entirely during training. Finally, we conducted experiments to determine whether a base or a post-trained model is a better starting point for fine-tuning. We found that base models tend to learn more effectively than their post-trained counterparts. Appendix~\ref{appendix:training_recipe_adds} presents additional training details, alongside a small-scale experiment comparing Qwen3-8B and Qwen3-8B-post-trained under the same training recipe, illustrating the differences between starting from a base versus a post-trained model.


\subsection{Evaluation benchmarks}
\label{eval_benchmarks}
We evaluate models exclusively on cybersecurity benchmarks spanning governance/compliance, architecture/operations, and threat detection \& response. The suite emphasizes document-grounded reasoning, consistent mapping across security taxonomies, and resilience to adversarial distractors.
Our evaluation uses the benchmarks listed below; See Appendix \ref{appen_eval_benchmarks} for further details.

\textbf{CTI-MCQ} tests breadth of CTI knowledge via multiple-choice items on attack patterns, actors, detections, mitigations, and frameworks \citep{ctibench}.
\textbf{CTI-RCM} evaluates document-grounded root-cause mapping by linking CVE evidence and bug reports to the correct CWE(s) with taxonomy-aware disambiguation \citep{ctibench}.
\textbf{SecEval} offers over 2,000 multi-option questions across nine security domains from authoritative sources, measuring accurate recall and the ability to apply controls and frameworks to concrete scenarios \citep{sec_eval}.
\textbf{CyberMetric-2000} comprises 2000 expert-validated questions spanning diverse subdomains, indicating professional-level declarative security knowledge under closed-book conditions \citep{tihanyi2024cybermetric}.
\textbf{CISSP exams} contains questions drawn from the assessment tests within the CISSP learning material, assessing analysts' skills across the entire security posture. 
\textbf{Technical Weakness Impact Mapping} requires assigning 
CWE descriptions 
a weakness to one or more of eight technical impacts, emphasizing consequence-centric reasoning beyond exploitability \citep{levi2025cyberpal}. 
\textbf{Adversarial CTI} ties questions to specific MITRE ATT\&CK entities and uses adversarial distractors to probe robustness on campaigns, tactics, detections, and mitigations \citep{levi2025cyberpal}.
\textbf{CTI Detection \& Mitigation} checks whether models propose appropriate detections and mitigations for tactics/techniques, attack patterns, weaknesses, and vulnerabilities \citep{levi2025cyberpal}.
\textbf{CTI Relationship Prediction} tests cross-taxonomy reasoning and relationship \textit{hallucinations} by choosing the correct justification for whether two CTI entities (e.g., CVE and CWE mapping) are related \citep{levi2025cyberpal}.

\begin{figure}[h]
  \centering
  \includegraphics[width=\columnwidth]{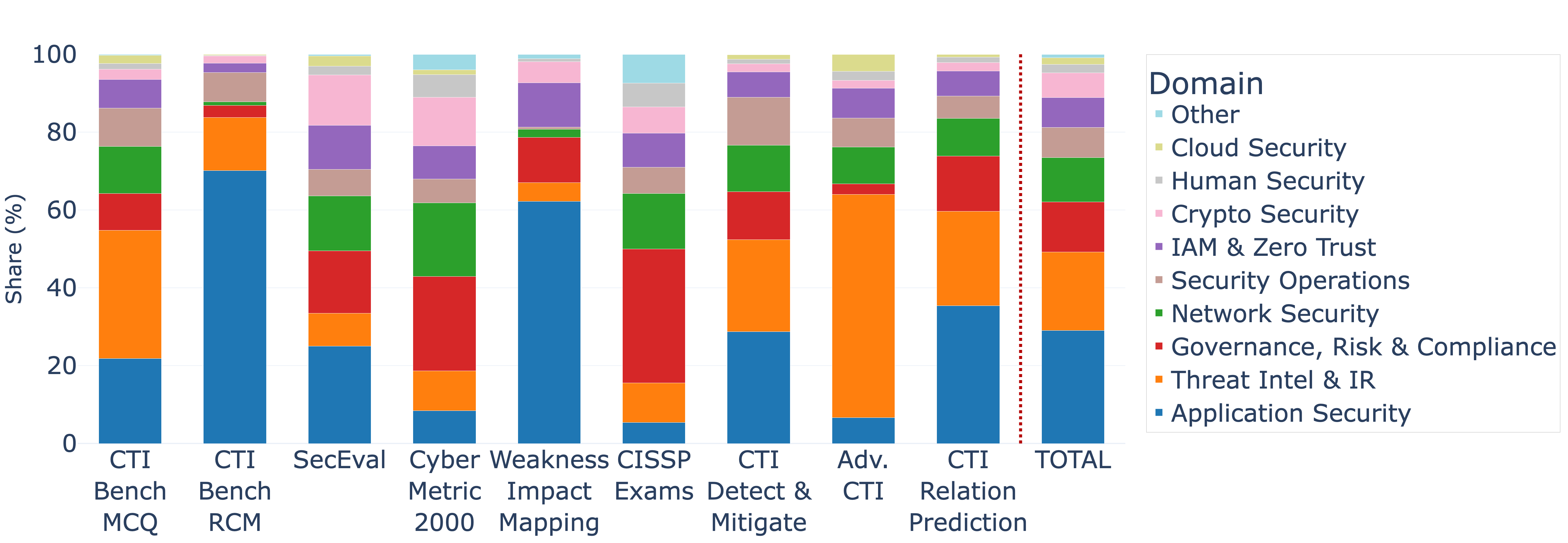}
  \caption{Domain composition of each data source in the evaluation benchmarks.}
  \label{figures:eval-stats}
\end{figure}

Figure~\ref{figures:eval-stats} shows the benchmark’s distribution across the cybersecurity taxonomy used to quantify coverage
(see Appendix~\ref{appendix:cyber_taxonomy_categories}).
As can be seen, the benchmarks are closely aligned with organizational security priorities, with a particular focus on \textit{threat Intelligence}, \textit{incident response}, \textit{security operations}, \textit{application security}, and \textit{identity management}.

\subsection{Evaluation Process and Metrics}
\label{eval_process}
Similar to \cite{wang2024mmlu}, we also found that models utilizing Chain of Thought (CoT) reasoning achieved better performance on complex security benchmarks, compared to direct answering. Therefore, we utilize a zero-shot CoT prompting. The CoT template incorporates essential reasoning steps and format to allow models to easily follow the given instructions. 
We used zero-temperature for consistency.
We then use a regular expression parser to extract the final answer from the model's CoT process. The prompt used in the evaluation is provided in Figure~\ref{figures:eval-prompt} in Appendix \ref{template_and_prompts}. For the Qwen suite of models, we compared our fine-tuned versions to baseline models (post-trained) with the \textit{thinking} flag enabled, allowing them to leverage their reasoning process. For gpt-oss, we used reasoning effort \textit{Medium} to avoid failures caused by the full CoT exceeding 
maximum 
window size.

\section{CyberPal~2.0: A Suite of Cybersecurity Language Models}
\label{results}
To demonstrate the effectiveness of our method, we train a family of security-expert SLMs ranging from 4B to 20B parameters. We then report results against the post-trained versions of the same base models from which our models were fine-tuned, alongside results against state-of-the-art frontier models (e.g., o1,  Sec-Gemini~v1). Lastly, we perform ablation studies.

\subsection{Results}

\textbf{Results Compared to Baselines.}
Table~\ref{table:res} presents results compared to the baseline models from the same families that CyberPal~2.0 was fine-tuned from. Meaning, for example, we measure improvements across various benchmarks between Qwen3-8B and CyberPal~2.0-8B, which was fine-tuned from Qwen3-8B-base. 
On average, our models outperform their baselines by 7--14\%. We also observe substantial gains on key benchmarks such as CTIBench-RCM, where our models exceed the baselines by 16--31\%, and CTIBench-MCQ, where they achieve improvements of 8--12\%.  
We also include gpt-oss-120B as a reference to highlight our models' strong performance.

\begin{table*}[t]
\caption{Evaluation results for CyberPal~2.0 models compared to their corresponding baseline (post-trained) models and the gpt-oss-120B open-source model.}
\label{table:res}
\begin{center}
\scalebox{0.85}{
\renewcommand{\arraystretch}{1.4} 
\begin{tabular}{l|ccccccccc|c}
\hline
Model &
  \multicolumn{1}{l}{\begin{tabular}[c]{@{}l@{}}CTI\\Bench \\ MCQ\end{tabular}} &
  \multicolumn{1}{l}{\begin{tabular}[c]{@{}l@{}}CTI\\Bench\\ RCM\end{tabular}} &
  \multicolumn{1}{l}{SecEval} &
  \multicolumn{1}{l}{\begin{tabular}[c]{@{}l@{}}Cyber\\ Metric\\ 2000\end{tabular}} &
  \multicolumn{1}{l}{\begin{tabular}[c]{@{}l@{}}CISSP\\ Exams\end{tabular}} &
  \multicolumn{1}{l}{\begin{tabular}[c]{@{}l@{}}Adv.\\ CTI\end{tabular}} &
  \multicolumn{1}{l}{\begin{tabular}[c]{@{}l@{}}Weakness\\ Impact \\ Mapping\end{tabular}} &
  \multicolumn{1}{l}{\begin{tabular}[c]{@{}l@{}}CTI\\ Detect \&\\ Mitigate\end{tabular}} &
  \multicolumn{1}{l|}{\begin{tabular}[c]{@{}l@{}}CTI\\ Relationship\\ Prediction\end{tabular}} &
  Avg. \\ \hline
Qwen3-4B &
  61.88 &
  49.95 &
  57.38 &
  87.40 &
  79.80 &
  64.51 &
  57.02 &
  60.77 &
  67.99 &
  65.19 \\
\textbf{CyberPal-2.0-4B} &
  \textbf{\begin{tabular}[c]{@{}c@{}}\gain{69.70}\\ \gain{(+7.82)}\end{tabular}} &
  \textbf{\begin{tabular}[c]{@{}c@{}}\gain{81.15}\\ \gain{(+31.20)}\end{tabular}} &
  \textbf{\begin{tabular}[c]{@{}c@{}}\gain{59.02}\\ \gain{(+1.64)}\end{tabular}} &
  \textbf{\begin{tabular}[c]{@{}c@{}}\gain{87.80}\\ \gain{(+0.40)}\end{tabular}} &
  \textbf{\begin{tabular}[c]{@{}c@{}}\gain{80.80}\\ \gain{(+1.00)}\end{tabular}} &
  \textbf{\begin{tabular}[c]{@{}c@{}}\gain{68.03}\\ \gain{(+3.52)}\end{tabular}} &
  \textbf{\begin{tabular}[c]{@{}c@{}}\gain{66.48}\\ \gain{(+9.46)}\end{tabular}} &
  \textbf{\begin{tabular}[c]{@{}c@{}}\gain{64.03}\\ \gain{(+3.26)}\end{tabular}} &
  \textbf{\begin{tabular}[c]{@{}c@{}}\gain{77.12}\\ \gain{(+9.13)}\end{tabular}} &
  \textbf{\begin{tabular}[c]{@{}c@{}}\gain{72.68}\\ \gain{(+7.49)}\end{tabular}} \\ \hline
Qwen3-8B &
  63.13 &
  63.25 &
  56.19 &
  88.45 &
  83.33 &
  64.93 &
  53.58 &
  59.88 &
  60.67 &
  65.93 \\
\textbf{CyberPal-2.0-8B} &
  \textbf{\begin{tabular}[c]{@{}c@{}}\gain{75.15}\\ \gain{(+12.02)}\end{tabular}} &
  \textbf{\begin{tabular}[c]{@{}c@{}}\gain{85.95}\\ \gain{(+22.70)}\end{tabular}} &
  \textbf{\begin{tabular}[c]{@{}c@{}}\gain{66.93}\\ \gain{(+10.74)}\end{tabular}} &
  \textbf{\begin{tabular}[c]{@{}c@{}}\gain{89.85}\\ \gain{(+1.40)}\end{tabular}} &
  \textbf{\begin{tabular}[c]{@{}c@{}}\gain{88.89}\\ \gain{(+5.56)}\end{tabular}} &
  \textbf{\begin{tabular}[c]{@{}c@{}}\gain{87.61}\\ \gain{(+22.68)}\end{tabular}} &
  \textbf{\begin{tabular}[c]{@{}c@{}}\gain{71.06}\\ \gain{(+17.48)}\end{tabular}} &
  \textbf{\begin{tabular}[c]{@{}c@{}}\gain{70.26}\\ \gain{(+10.38)}\end{tabular}} &
  \textbf{\begin{tabular}[c]{@{}c@{}}\gain{87.66}\\ \gain{(+26.99)}\end{tabular}} &
  \textbf{\begin{tabular}[c]{@{}c@{}}\gain{80.37}\\ \gain{(+14.44)}\end{tabular}} \\ \hline
Qwen3-14B &
  64.28 &
  70.50 &
  61.48 &
  89.85 &
  86.36 &
  69.43 &
  62.46 &
  63.44 &
  58.48 &
  69.59 \\
\textbf{CyberPal-2.0-14B} &
  \textbf{\begin{tabular}[c]{@{}c@{}}\gain{75.51}\\ \gain{(+11.23)}\end{tabular}} &
  \textbf{\begin{tabular}[c]{@{}c@{}}\gain{86.00}\\ \gain{(+15.50)}\end{tabular}} &
  \textbf{\begin{tabular}[c]{@{}c@{}}\gain{69.71}\\ \gain{(+8.23)}\end{tabular}} &
  \textbf{\begin{tabular}[c]{@{}c@{}}\gain{89.95}\\ \gain{(+0.10)}\end{tabular}} &
  \textbf{\begin{tabular}[c]{@{}c@{}}\gain{90.40}\\ \gain{(+4.04)}\end{tabular}} &
  \textbf{\begin{tabular}[c]{@{}c@{}}\gain{89.58}\\ \gain{(+20.15)}\end{tabular}} &
  \textbf{\begin{tabular}[c]{@{}c@{}}\gain{70.77}\\ \gain{(+8.31)}\end{tabular}} &
  \textbf{\begin{tabular}[c]{@{}c@{}}\gain{70.95}\\ \gain{(+7.51)}\end{tabular}} &
  \textbf{\begin{tabular}[c]{@{}c@{}}\gain{92.93}\\ \gain{(+34.45)}\end{tabular}} &
  \textbf{\begin{tabular}[c]{@{}c@{}}\gain{81.76}\\ \gain{(+12.17)}\end{tabular}} \\ \hline
gpt-oss-20B &
  64.57 &
  68.95 &
  67.65 &
  \textbf{90.20} &
  79.80 &
  61.83 &
  \textbf{71.91} &
  67.49 &
  65.42 &
  70.87 \\
\textbf{CyberPal-2.0-20B} &
  \textbf{\begin{tabular}[c]{@{}c@{}}\gain{75.71}\\ \gain{(+11.14)}\end{tabular}} &
  \textbf{\begin{tabular}[c]{@{}c@{}}\gain{87.40}\\ \gain{(+18.45)}\end{tabular}} &
  \textbf{\begin{tabular}[c]{@{}c@{}}\gain{72.86}\\ \gain{(+5.21)}\end{tabular}} &
  89.05 &
  \textbf{\begin{tabular}[c]{@{}c@{}}\gain{86.87}\\ \gain{(+7.07)}\end{tabular}} &
  \textbf{\begin{tabular}[c]{@{}c@{}}\gain{84.93}\\ \gain{(+23.10)}\end{tabular}} &
  70.77 &
  \textbf{\begin{tabular}[c]{@{}c@{}}\gain{67.69}\\ \gain{(+0.20)}\end{tabular}} &
  \textbf{\begin{tabular}[c]{@{}c@{}}\gain{87.66}\\ \gain{(+22.24)}\end{tabular}} &
  \textbf{\begin{tabular}[c]{@{}c@{}}\gain{80.33}\\ \gain{(+9.46)}\end{tabular}} \\ \hline
gpt-oss-120B &
  69.37 &
  79.95 &
  68.02 &
  92.55 &
  84.34 &
  72.76 &
  65.90 &
  64.52 &
  70.56 &
  74.21 \\ \hline
\end{tabular}
}
\end{center}
\end{table*}



\textbf{Results Compared to Open Source cybersecurity models \& Additional Open Source Model Families.} 
We evaluated our models against recent 7B–8B open-source cybersecurity models; our 8B  leads across all benchmarks. Full details are in Appendix \ref{appendix:open_source_models_performance}. Additionally, in Appendix \ref{app-comp-other-model-families} we evaluate our models against other open source model families.


\textbf{Results Compared to Frontier LLMs.}
We evaluated our models against state-of-the-art general models such as Sec-Gemini v1 and OpenAI's o1. As evaluation is costly, we follow Sec-Gemini~v1 evaluation protocol and report results on the CTIBench benchmarks: CTIBench-MCQ and CTIBench-RCM \citep{ctibench}. 
CTIBench-MCQ assesses an LLM’s understanding of core cyber–threat intelligence concepts, while CTIBench-RCM evaluates model’s ability to perform Root Cause Mapping (RCM), identifying the underlying causes of vulnerabilities by correlating vulnerabilities records and bug tickets with weaknesses. This benchmark is considered a leading threat intelligence benchmark and serves as a strong indicator of a model's threat management capabilities. 

As shown in Figure~\ref{fig:sota}, our models are on par with—or better than—most frontier models. \textbf{RCM:} CyberPal~2.0–20B ranks first overall, surpassing Sec-Gemini~v1; the 14B, 8B, and 4B variants all ranked second and exceed the remaining frontier models.
\textbf{MCQ:} the 20B and 14B models ranks second and third respectively, immediately behind Sec-Gemini~v1 and ahead of o1; the 8B model is competitive with GPT-4o; and even the 4B model outperforms much larger models such as Mistral Large and DeepSeek-v3, with performance close to o3-mini.


\subsection{Results on Real-World Use Cases}
\label{real_world_use_cases_main}

Beyond standard cybersecurity benchmarks, we evaluate CyberPal~2.0 on two applied workflows: mapping threat reports to MITRE ATT\&CK techniques and reassessing CVE severity under product-specific constraints. These tasks test whether the models generalize to realistic security artifacts and operational decision-making settings. Additional real-world evaluations, including CyberSOCEval and secure code generation, are reported in Appendix~\ref{app-real-world-use-cases}.

\paragraph{Threat reports to TTP mapping.}
We construct this benchmark from MITRE campaign entries and their references. Each campaign lists techniques used by the adversary, and some technique entries cite APT reports describing the use of that technique. We use these reports to build a multiple-choice task in which the model receives an unstructured threat report and selects the corresponding ATT\&CK technique. This evaluates whether the model can connect realistic CTI evidence to operational security taxonomies.

\paragraph{CVE reassessment.}
We also evaluate CVE reassessment, where the model must reassess a CVSS\footnote{https://www.first.org/cvss/v3.1/specification-document} score for a specific package, product, or deployment context. This setting tests whether the model can adjust generic vulnerability severity according to concrete deployment constraints such as packaging, exposure, configuration, and permissions. We report normalized Mean Average Distance, where higher is better.

Table~\ref{table:real_world_main} shows consistent gains across both settings. On threat-report mapping, CyberPal~2.0 improves over all corresponding baselines: all models gain 2.75--8.75 points. On CVE reassessment, all CyberPal~2.0 variants improve over their baselines, with the largest gains for the 14B and 20B models. These results suggest that SecKnowledge~2.0 improves practical security reasoning beyond standard benchmark formats.

\begin{table}[!t]
    \caption{Real-world use-case results.}
    \label{table:real_world_main}
    \centering
    \small
    \setlength{\tabcolsep}{6pt}
    \renewcommand{\arraystretch}{1.15}
    \begin{tabular}{l c c}
        \toprule
        \textbf{Model} &
        \textbf{\begin{tabular}[c]{@{}c@{}}Threat$\rightarrow$TTP\\(Accuracy)\end{tabular}} &
        \textbf{\begin{tabular}[c]{@{}c@{}}CVE Reassess.\\(Norm. MAD)\end{tabular}} \\
        \midrule
        Qwen3-4B & 74.00 & 0.783 \\
        \textbf{CyberPal~2.0-4B} &
        \textbf{\begin{tabular}[c]{@{}c@{}}\gain{82.75}\\ \gain{(+8.75)}\end{tabular}} &
        \textbf{\begin{tabular}[c]{@{}c@{}}\gain{0.830}\\ \gain{(+0.047)}\end{tabular}} \\
        \midrule
        Qwen3-8B & 74.00 & 0.825 \\
        \textbf{CyberPal~2.0-8B} &
        \textbf{\begin{tabular}[c]{@{}c@{}}\gain{78.25}\\ \gain{(+4.25)}\end{tabular}} &
        \textbf{\begin{tabular}[c]{@{}c@{}}\gain{0.834}\\ \gain{(+0.009)}\end{tabular}} \\
        \midrule
        Qwen3-14B & 79.25 & 0.740 \\
        \textbf{CyberPal~2.0-14B} &
        \textbf{\begin{tabular}[c]{@{}c@{}}\gain{85.25}\\ \gain{(+6.00)}\end{tabular}} &
        \textbf{\begin{tabular}[c]{@{}c@{}}\gain{0.833}\\ \gain{(+0.093)}\end{tabular}} \\
        \midrule
        gpt-oss-20B & 76.75 & 0.738 \\
        \textbf{CyberPal~2.0-20B} &
        \textbf{\begin{tabular}[c]{@{}c@{}}\gain{79.50}\\ \gain{(+2.75)}\end{tabular}} &
        \textbf{\begin{tabular}[c]{@{}c@{}}\gain{0.834}\\ \gain{(+0.096)}\end{tabular}} \\
        \bottomrule
    \end{tabular}

    \vspace{-0.8cm}
    \footnotesize
    \raggedright
\end{table}

\subsection{Ablation Studies}
\label{ablation_sec}
We conduct ablation studies to identify which components of SecKnowledge 2.0 are responsible for the observed gains. In particular, we study three factors: (1) the data pipeline used to construct the training set, (2) the contribution of retrieval grounding through the search component, and (3) the choice of reformatting backbone LLM. All ablations are run with the same target model (Qwen3-4B-Base), training budget, configuration, optimization settings, and context settings.(as described in \S\ref{train_recipe}) The ablation were further evaluated on all nine benchmarks. Table~\ref{tab:ablation_summary} summarizes the average accuracy across all ablations, and Appendix~\ref{appendix:ablation} reports the full benchmark-level results.

\begin{table*}[t]
\caption{Summary of ablations on the Qwen3-4B-base setting. We report average accuracy across all nine benchmarks.}
\label{tab:ablation_summary}
\centering
\small
\setlength{\tabcolsep}{6pt}
\renewcommand{\arraystretch}{1.1}
\renewcommand{\tabularxcolumn}[1]{m{#1}}
\begin{tabularx}{0.7\textwidth}{Y Y C c c}
\toprule
\textbf{Data Pipeline} & \textbf{Formats} & \textbf{Backbone LLM} & \textbf{Search} & \textbf{Avg.} \\
\midrule
Qwen3-4B base & -- & -- & -- & 65.19 \\
\cmidrule(lr){1-5}
SecKnowledge \citep{levi2025cyberpal} & -- & -- & -- &
\begin{tabular}[c]{@{}c@{}}
\gain{68.60} \\
\gain{(+3.41)}
\end{tabular} \\
\cmidrule(lr){1-5}
SecKnowledge 2.0 (Ours) &
\begin{tabular}[c]{@{}l@{}}General-purpose\\(\citet{fan2024reformatted})\end{tabular} &
gpt-oss-120B & \ding{51} &
\begin{tabular}[c]{@{}c@{}}
\gain{70.04} \\
\gain{(+4.85)}
\end{tabular} \\
\cmidrule(lr){1-5}
SecKnowledge 2.0 (Ours) & Expert-written & gpt-oss-120B & \ding{55} &
\begin{tabular}[c]{@{}c@{}}
\gain{71.11} \\
\gain{(+5.92)}
\end{tabular} \\
\cmidrule(lr){1-5}
SecKnowledge 2.0 (Ours) & Expert-written & Llama-4-Maverick & \ding{51} &
\begin{tabular}[c]{@{}c@{}}
\gain{72.32} \\
\gain{(+7.13)}
\end{tabular} \\
\cmidrule(lr){1-5}
SecKnowledge 2.0 (Ours) & Expert-written & gpt-oss-120B & \ding{51} &
\textbf{\begin{tabular}[c]{@{}c@{}}
\gain{72.68} \\
\gain{(+7.49)}
\end{tabular}} \\
\bottomrule
\end{tabularx}
\end{table*}

\paragraph{Effect of the data pipeline.}

We first compare models trained on the original SecKnowledge dataset with models trained on data produced by a reformatting baseline inspired by \citet{fan2024reformatted}, and with models trained on data produced by our full SecKnowledge 2.0 pipeline. Specifically, in the baseline setting we keep our pipeline fixed and rewrite the original SecKnowledge dataset using general-purpose question-answering formats instead of expert-written task-specific formats. Training on the original SecKnowledge already improves over the base model, confirming the value of domain-specific cybersecurity supervision. Applying generic reformatting yields further gains, but our full SecKnowledge 2.0 pipeline performs best overall. This indicates that the improvements are not explained by reformatting alone, but by using expert-defined task structures and enriched supervision tailored to the cybersecurity domain.

\paragraph{Effect of retrieval grounding.}

We next assess the contribution of the search component by comparing the full pipeline with a variant in which search is removed. Removing search reduced the average performance, showing that retrieval-based grounding improves the quality of the generated supervision. This result is consistent with our hypothesis that external evidence helps reduce hallucinations and produces more reliable rewritten answers.

\paragraph{Effect of the reformatting backbone LLM.}

Finally, we examine the effect of replacing the reformatting backbone LLM from gpt-oss-120B with Llama-4-Maverick \citep{meta_llama4_2025} while keeping the rest of the pipeline unchanged. Despite the large difference in scale between the two reformatters (120B vs. 400B parameters), the resulting average performance is nearly unchanged. This suggests that the gains are not tied to a specific large teacher model, and that a different reformatter can support almost the same downstream performance within our pipeline.

\subsection{LLM-as-a-Judge Evaluation}
Finally, we assess answer quality via \emph{LLM-as-a-Judge} (LLMaaJ) \citep{zheng2023judging}. Thirty cybersecurity experts authored 115 open-ended questions spanning command-line risk assessment, enterprise security, general cybersecurity, network security, and CTI-related topics. For each question and pair of model answers, the judge receives expert-curated grounding documents.
We use OpenAI's o3 as the judge.

We evaluate answer quality using pairwise LLM-as-a-Judge comparisons. To validate the judge, we measure its agreement with human cybersecurity experts. We find that o3 agrees with human preferences in 80\% of cases without grounding, and in over 90\% of cases when provided with carefully collected grounding documents. This suggests that grounded o3 judgments are a reliable proxy for human preference in our setting.

\begin{figure}[h]
  \centering
  \includegraphics[width=0.8\columnwidth]{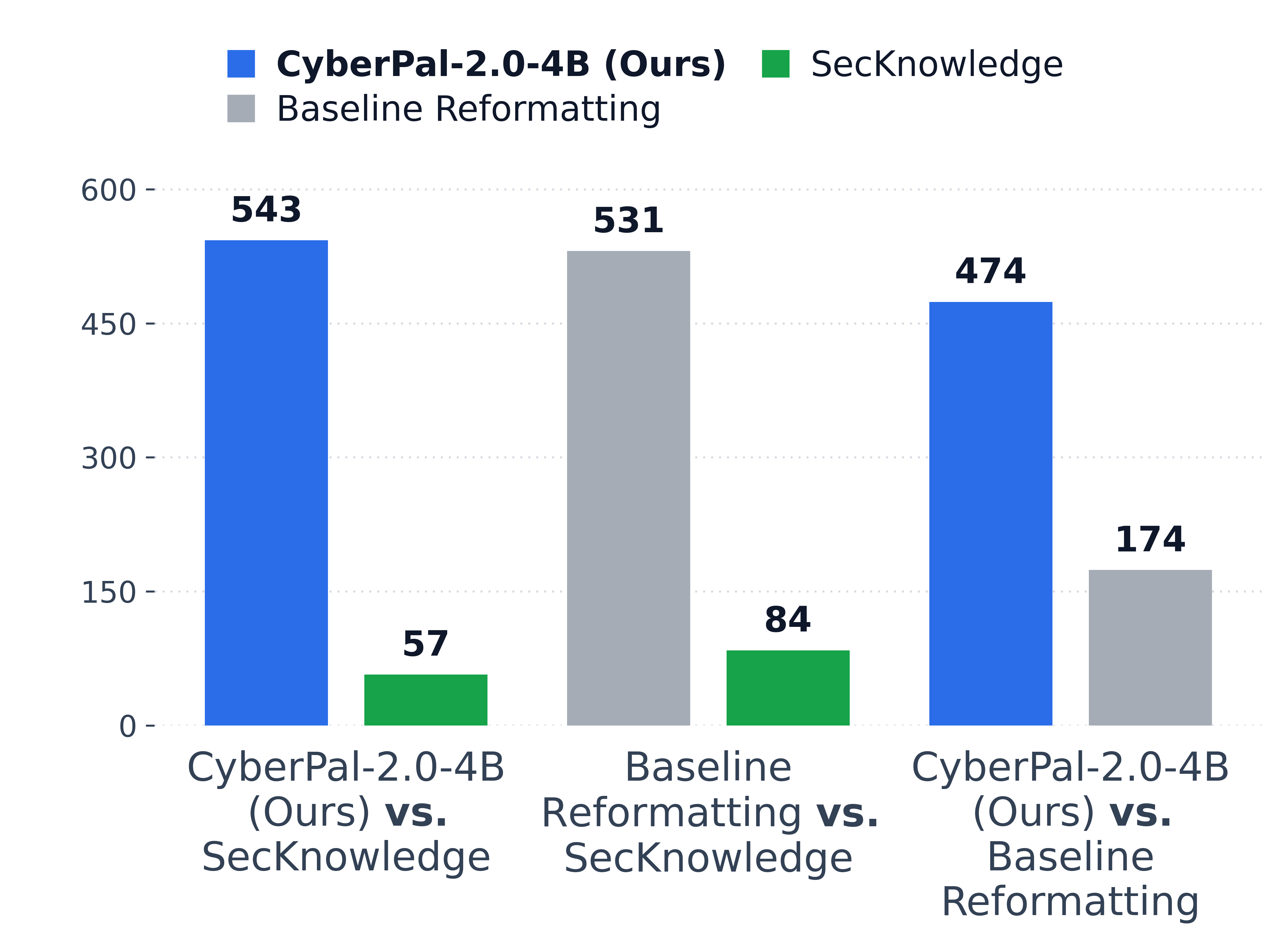}
  \caption{Pairwise comparison results from LLMaaJ (o3) with \textit{grounding}, aligned with human experts.}
  \label{fig:ablation:llmaj}
\end{figure}

Using the previously constructed open-ended cybersecurity questions, we generate answers from different model variants and ask the judge to compare each pair of responses. The judge receives the question, two candidate answers, and the relevant grounding documents, and decides which answer is better. Figure~\ref{fig:ablation:llmaj} shows that CyberPal-2.0 is consistently preferred over both baselines. Further details on the question set, evaluation protocol, and additional per-category results are provided in Appendix~\ref{appendix:llmaaj}.

\section{Conclusion}
\label{conclusion}
\textbf{CyberPal~2.0} demonstrates that compact, domain-specific SLMs (4B–20B) can deliver frontier-level capability for security operations without frontier-level cost. Built on \textit{SecKnowledge~2.0}—with schema-driven reformatting, expert-in-the-loop enrichment, and a multi-step grounding process—our models achieve 7--14\% average gains over strong open-source baselines and, on core CTI tasks, match or surpass leading closed models; 
notably, the 20B model outperforms GPT-4o, o1, o3-mini, and Sec-Gemini~v1, while even the 4B variant ranks second. Ablations and LLMaaJ validated by human experts attribute the gains primarily to data-quality improvements from our \textbf{SecKnowledge~2.0} enhanced reformatting and enrichment pipeline.

\newpage

\section*{Impact Statement}
While our training focused on defensive threat management and security operations (e.g., threat investigation and incident response), the models’ enhanced security knowledge could be misused by malicious actors in unforeseen ways. To reduce misuse risk, we (i) rely mostly on publicly available, open-access, non-sensitive sources; (ii) avoid training or releasing offensive cybersecurity capabilities; and (iii) will distribute models under responsible-use terms with safety filters and red-teaming.

\bibliography{example_paper}
\bibliographystyle{icml2026}

\newpage
\appendix
\onecolumn

\section{Pipeline Extensions Details}
\label{appendix:pipeline_extentions}

\subsection{Format Generation Framework}
\label{format_generation_framework}

\begin{figure}[h!]
  \centering
  \includegraphics[width=\textwidth, height=0.72\paperheight]
  {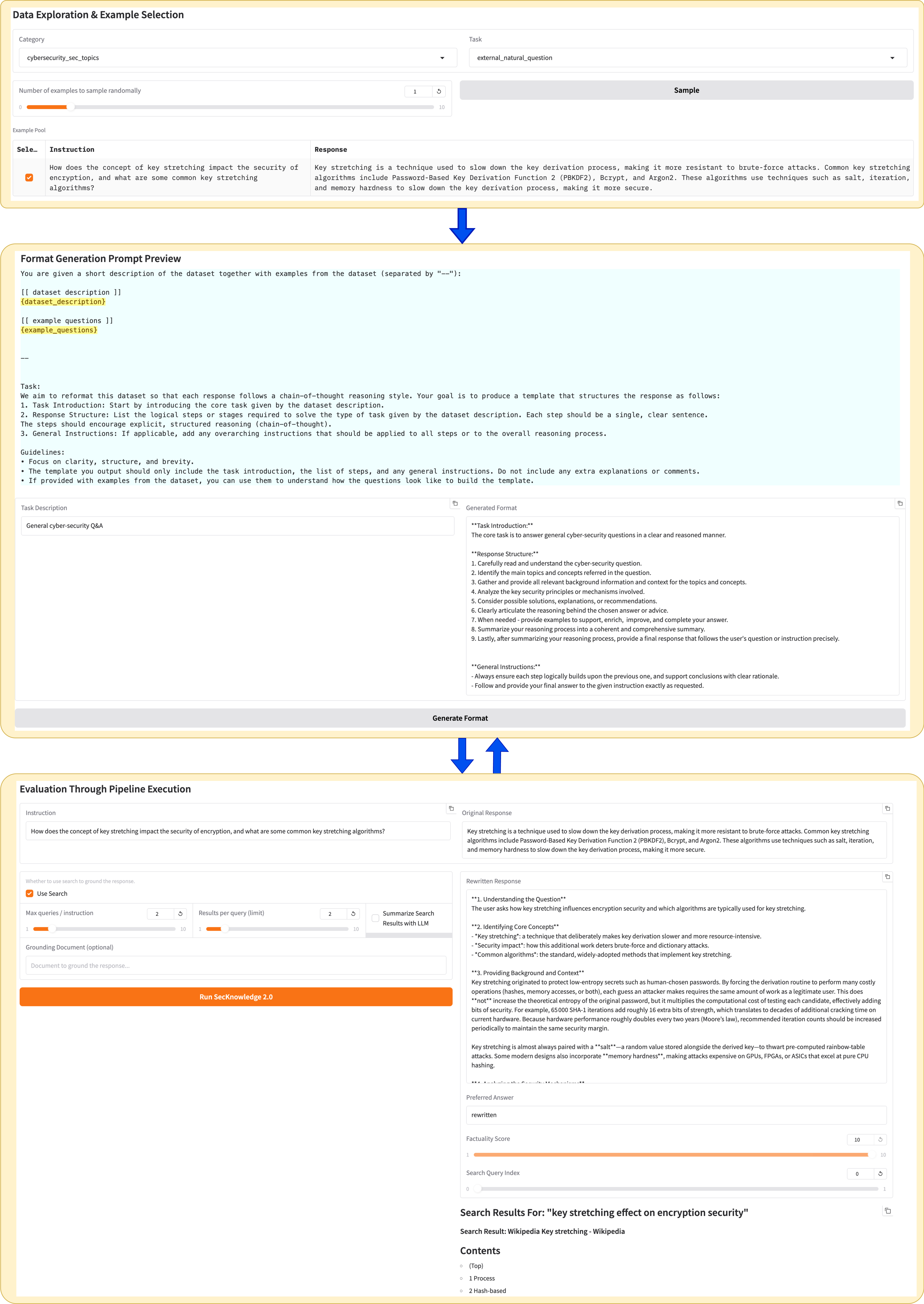}
  \caption[Format generation UI screenshots]{Screenshots from the UI designed for our format generation framework. First, the relevant task is selected, and random examples are sampled from the task data partition, then a candidate format is generated from them using a LLM and the prompt in blue, afterwards an expert can configure and run the pipeline, and edit the format if needed. The UI was developed using Gradio\protect\footnotemark.}
  \label{figures:format-generation-ui}
\end{figure}

The system is composed of three primary stages, with the second and third stages forming an iterative loop that can be repeated until the user is satisfied with the resulting format, as illustrated in Figure~\ref{figures:format-generation-ui}. It is important to note that this framework is applicable only once the set of tasks covering the entire dataset has been defined, at which point the process is limited to the generation of formats.

\textbf{Data Exploration \& Example Selection.} First, the user selects the appropriate category and task from the menus. From \(N\) available examples for each task (\(N=500\) in our case), the system samples \(k\) instruction–answer pairs for inspection (\(k=1\) in Fig. \ref{figures:format-generation-ui}). This stage enables the user to explore the dataset - examining the range of questions and corresponding answers for the selected task - and to identify representative examples. These examples are subsequently used both to guide format generation and as inputs to the pipeline.

\textbf{Format Generation.} Second, the user provides a brief description of the task, this description will be used both to classify unlabeled instructions from the dataset and to generate the format. Then he selects an LLM to produce a candidate format using one of the available prompts. In our case, two distinct prompts were required: one tailored for specific tasks - such as the instructions generated in the first stage of \textit{SecKnowledge}, which originate from a defined source and consistently ask for the same type of information, albeit in different contexts - and another designed for more general tasks, which encompass a wide variety of instructions, as in the second stage of \textit{SecKnowledge}. In the latter case, providing examples may bias the format toward the selected instances, which is undesirable. The framework further supports the seamless addition of new prompts if needed. Once generated, the format can be refined by the user through manual editing.

\textbf{Evaluation Through Pipeline Execution.} Third, the user can run the pipeline on any example, with the first example automatically pre-filled by default. During this step, the user may adjust various hyper-parameters - for instance, enabling or disabling web search, specifying the number of search queries and the number of results per query, and deciding whether to summarize each retrieved before including it in the rewriting context. A grounding document can also be provided, either as an alternative to or in addition to web search. The pipeline then outputs the rewritten response, quality assessment scores, and, if requested, the retrieved search results.

\footnotetext{{https://www.gradio.app/}}

\subsection{Data Generation Quality Assessment}
\label{pipeline_quality_assessment}

After rewriting the original answer according to the format, the pipeline also incorporates evaluation in the form of LLM as a Judge. There are 2 criteria by which we judge the answers generated by the pipeline.

\begin{enumerate}
    \item{\textbf{Readability.} We prompt the judge with the instruction, and both answers and ask it to select the better answer according to the criteria described in Figure~\ref{figures:llmaaj-prompt} (original and rewritten). We do this two times - in the first time the original answer is first and the rewritten is second, and in the second time the order is the opposite (while anonymizing which one is the original one and which one is the rewritten one). We test both directions to avoid positional bias \citep{wang2023large, zheng2023judging}.}
    \item{\textbf{Factuality.} We prompt the judge with the original answer and the rewritten answer, emphasizing that the original answer is the ground truth, and ask the LLM to provide a score in a scale of 1-10 that determines how factual the rewritten answer with respect to the original answer.}
\end{enumerate}

\begin{figure}[h]
  \centering
  \includegraphics[scale=0.32]
  {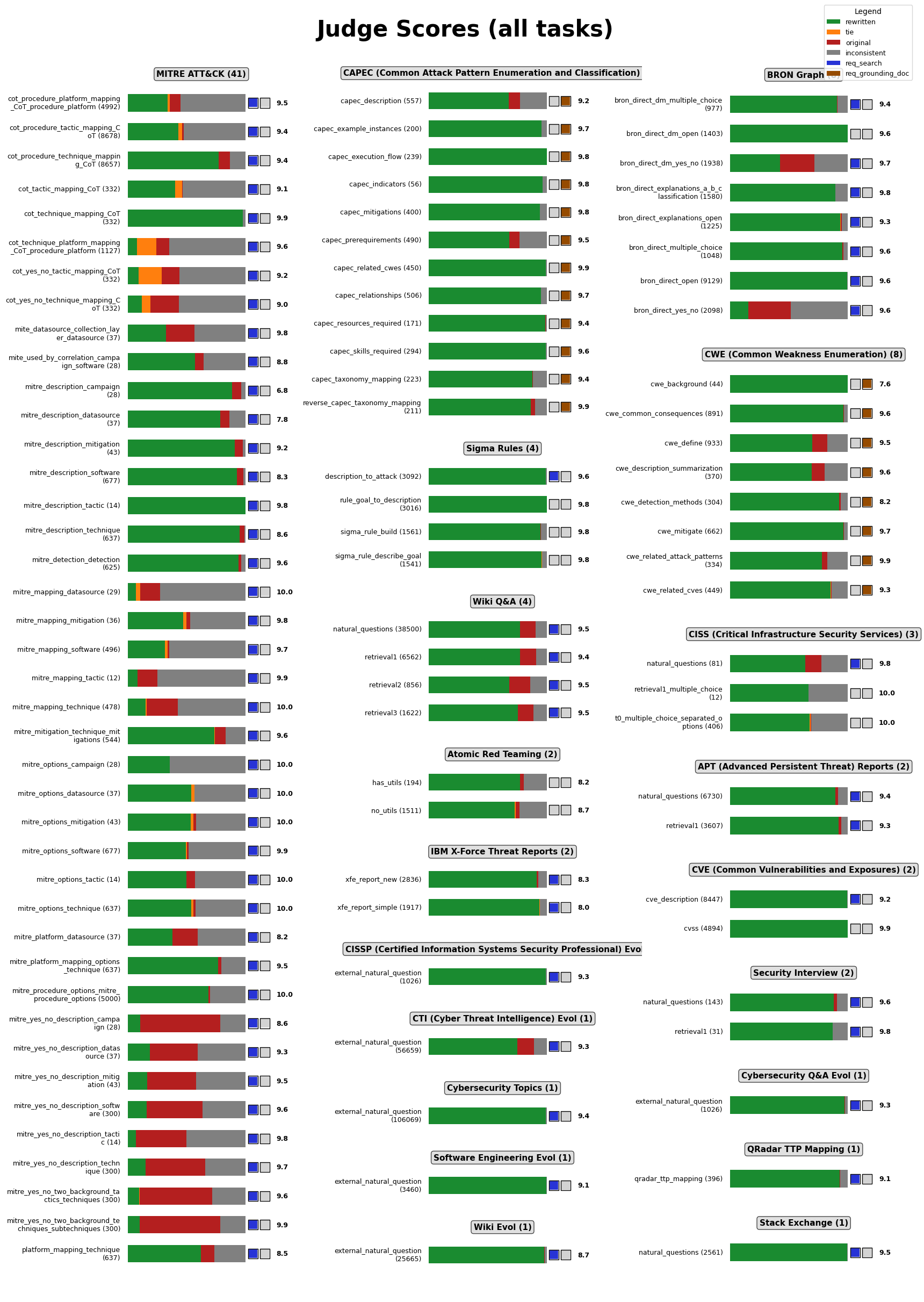}
  \caption{Data generation quality assessment scores for \textit{SecKnowledge 2.0}, broken down by task. Each bar is color-coded to indicate \textbf{readability} outcomes (green for rewritten, red for original, orange for tie, and gray for position inconsistency). The boxes to the right of each bar shows the context requirements (blue box for a task that requires web search, brown box for a task that requires a grounding document), and the number to the right of those boxes denotes the average \textbf{factuality} score. The framed labels above groups of tasks indicates their parent category (categories ending with "Evol" contain examples synthetically generated by the second phase of \textit{SecKnowledge}). The number in parentheses after each category indicates how many tasks belong to that category, while the number in parentheses after each task represents the count of instructions within that task.}
  \label{figures:llmaaj-secknowledge2}
\end{figure}

By combining these two criteria, we can get a sense of the quality of the defined formats and the new dataset. In Figure~\ref{figures:llmaaj-secknowledge2}, we present the quality assessment results on our dataset, \textit{SecKnowledge 2.0}. On average, the new answers are preferred 85.62\% of the time (with 5.55\% to the favor of the original answers, and 8.56\% of inconsistency, where switching the position of the answers changed the judge decision, the rest are ties), while maintaining the factuality, reflected by an average factuality score of \(9.25\). These results show that our pipeline is robust.

\begin{figure*}[h]
  \centering
  \includegraphics[width=0.9\textwidth]
  {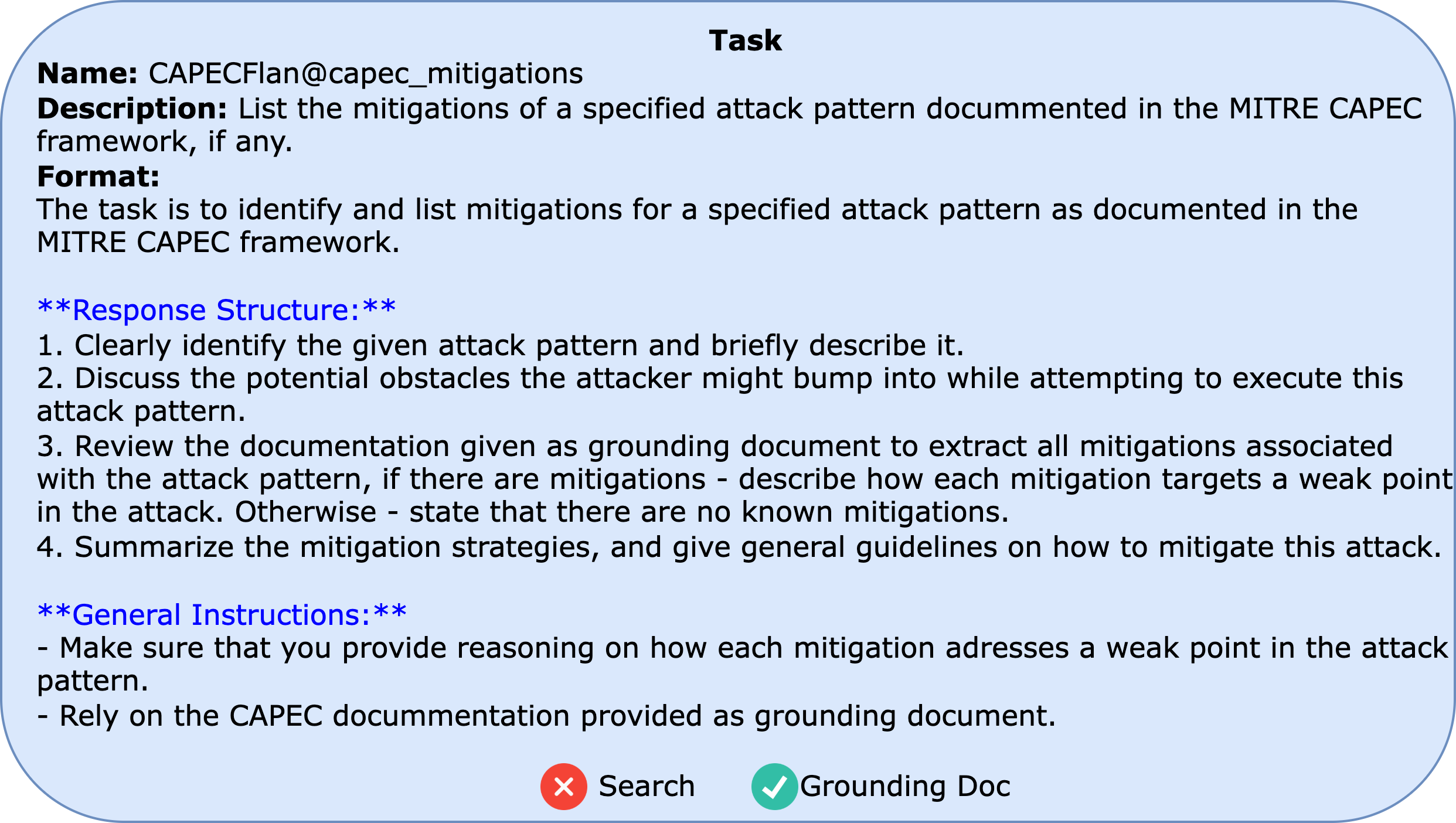}
  \caption{An example task, specifically the one illustrated in Figure~\ref{figures:realign-example}. A task consists of a name, a description, a format, whether it requires search, and whether it requires a grounding document.}
  \label{figures:format-example}
\end{figure*}

\section{Training Recipe Additional Details}
\label{appendix:training_recipe_adds}
\textbf{Training from base vs.\ post-trained model.}
In Table~\ref{tables:inst_vs_base}, we present a small-scale experiment examining the effect of the starting checkpoint \textit{using an identical training recipe and evaluation protocol}. Using Qwen3-8B, we evaluate how fine-tuning with \textit{SecKnowledge~2.0} affects performance when starting from the base model versus a post-trained model. We observed an interesting phenomenon: directly fine-tuning the base model (i.e., Qwen3-8B-Base) yields significantly better results than relying on the post-trained model (Qwen3-8B) as the starting point. This effect is amplified on benchmarks that require additional reasoning to arrive at the final answer (e.g., CTIBench-RCM). On average, fine-tuning from the base model provides a 15.16\% improvement across the key benchmarks, whereas starting from the post-trained model provides a 5.62\% improvement relative to the Qwen3-8B baseline—corresponding to a $2.7\times$ larger gain when initializing from the base model. 
For open-ended benchmarks such as CTIBench-RCM, the difference is even more pronounced: the model fine-tuned from the base checkpoint achieves a 22.7\% improvement, compared with 1.85\% for the model fine-tuned from the post-trained checkpoint. Although limited in scope, this experiment empirically indicates that, given sufficiently high data quality, initializing from a base model enables more effective learning than starting from a post-trained checkpoint that has already undergone extensive supervised fine-tuning and alignment--yet further work is needed to systematically disentangle how data quality and data scale interact with the choice of starting checkpoint during fine-tuning.

\begin{table}[h]
\caption{Comparing the improvement of fine-tuning our models when starting from base model vs. a post-trained model.}
\label{tables:inst_vs_base}
\begin{center}
\scalebox{0.9}{
\renewcommand{\arraystretch}{1.4} 
\begin{tabular}{l|ccc|c}
\hline
Model &
  \multicolumn{1}{l}{\begin{tabular}[c]{@{}l@{}}CTIBench\\ MCQ\end{tabular}} &
  \multicolumn{1}{l}{\begin{tabular}[c]{@{}l@{}}CTIBench\\ RCM\end{tabular}} &
  \multicolumn{1}{l|}{SecEval} &
  Avg. \\ \hline
Qwen 3 8b                                                                        & 63.13 & 63.25 & 56.19 & 60.85 \\
\begin{tabular}[c]{@{}l@{}}CyberPal2.0-8B (trained from Qwen3-8B)\end{tabular} & 68.90 & 65.10 & 65.42 & 66.47 \\
\begin{tabular}[c]{@{}l@{}}CyberPal2.0-8B (trained from Qwen3-8B-Base)\end{tabular} &
  \textbf{75.15} &
  \textbf{85.95} &
  \textbf{66.93} &
  \textbf{76.01} \\ \hline
\end{tabular}
}
\end{center}
\end{table}

\textbf{Incremental training methodology.} Lastly, consistent with the observations of \citet{mitra2023orca, levi2025cyberpal}, we empirically find that exposing the model to instructions of progressively increasing length—often correlated with task difficulty—enhances its learning capacity. Building on this principle, we adopt an incremental training methodology organized at the dataset level. Specifically, we first present the model with instructions from the original SecKnowledge dataset, followed by instructions from our new SecKnowledge~2.0 dataset.  

\textbf{Additional training details.} We select the final checkpoint by validation loss on a held-out split extracted from the training set. We train for two epochs, as we observe diminishing returns and an increased risk of overfitting thereafter. Training was conducted on a cluster of 12 NVIDIA A100 80\,GB GPU nodes, and evaluation was performed on NVIDIA H100 80GB GPUs.

\section{Evaluation benchmarks, statistics, and analysis}
\label{appendix:eval_sec}

\subsection{Evaluation benchmarks}
\label{appen_eval_benchmarks}

\textbf{CTI-MCQ} \citep{ctibench} is a multiple choice question benchmark aimed at assessing LLMs' capabilities in understanding crucial cyber threat intelligence concepts including  attack patterns, threat actors, APT campaigns, detection methods, mitigation strategies, common software vulnerabilities, attack pattern enumeration, alongside public CTI quizzes. 
This benchmark assess the breadth of CTI/domain knowledge; knowing frameworks/controls and when to apply them.

\textbf{CTI-RCM} CTI Root Cause Mapping (RCM) \citep{ctibench} identifies the underlying weakness(es) of a vulnerability by correlating CVE records and related bug tickets with CWE entries. Accurate root cause mapping is essential for guiding investments, policies, and practices aimed at addressing and eliminating these vulnerabilities. Strong LLM performance on CTI-RCM indicates grounded, document-linked reasoning and consistent, taxonomy-aware disambiguation—mapping real-world vulnerability evidence to the appropriate CWE(s) rather than relying on superficial keyword matches.

\textbf{SecEval} SecEval \citep{sec_eval} is a multiple-choice, multiple-option benchmark for evaluating LLMs’ cybersecurity knowledge, with over 2,000 questions spanning nine domains. 
SecEval was constructed using OpenAI GPT-4 from authoritative sources (open-licensed textbooks, official platform security docs, OWASP guides, CWE, and MITRE ATT\&CK/D3fend).
This benchmark assesses the breadth and accuracy of security/domain knowledge and the ability to choose and apply the right frameworks, controls, detections, and mitigation to concrete scenarios.


\textbf{CyberMetric 2000} CyberMetric \citep{tihanyi2024cybermetric} is a benchmark dataset for evaluating LLMs' knowledge in cybersecurity. The questions for the benchmark were created through a collaborative process, i.e., merging expert knowledge with LLMs. We used the 2000 questions dataset, verified by human evaluators, which covers a wide range of topics within cyber-security, validated by security experts.
As questions come from standards-grounded material and were validated by certified practitioners (e.g., CISSP/CISM/OSCP), strong performance primarily evidences professional-level declarative cybersecurity knowledge—accurate recall of definitions, controls, and best practices across diverse subdomains, and the ability to reject plausible distractors under closed-book conditions.



\textbf{CISSP Exams}
Introduced by \cite{levi2025cyberpal}, this benchmark uses exam-style questions from CISSP preparation materials to assess broad, professional cybersecurity knowledge across governance and risk, security architecture, operations, software and network security, and identity and access management. Items use plausible distractors and test principled reasoning and terminology rather than tool-specific tricks. A high score indicates strong declarative understanding, standards-aligned judgment, and the ability to separate best practices from common misconceptions under test conditions.

\textbf{Technical Weakness Impact Mapping} In CWE, each weakness, if successfully exploited, can lead to one or more technical impacts out of eight options: modify data, read data, DoS: unreliable execution, DoS: resource consumption, execute unauthorized code or commands, gain privileges / assume identity, bypass protection mechanism, and hide activities. This evaluation benchmark, introduced by \cite{levi2025cyberpal}, presents the model with CWEs and their descriptions, where the goal is to map each CWE to its related technical impact.
A high score indicates taxonomy-aware understanding of how specific weakness patterns translate into concrete consequences, beyond surface keyword matching. Because a single CWE can map to multiple impacts and descriptions are often terse, the benchmark primarily measures consequence reasoning rather than exploit feasibility or business risk. It thus serves as an impact-from-weakness signal that complements severity or exploitability evaluations.

\textbf{Adversarial CTI} \citet{levi2025cyberpal} compiled an adversarial evaluation dataset from various MITRE ATT\&CK sources to evaluate models on malicious software, campaigns, tactics, and corresponding detections and mitigations. Each input provides a question related to a specific MITRE instance, with the correct label being its corresponding source. To further challenge the models and test robustness, they introduced a novel adversarial attack for multiple-choice questions, where the attack chooses the false options that will confuse the model with the highest probability.

\textbf{CTI Detection and Mitigation} Introduced by \citet{levi2025cyberpal}, this benchmark is designed to assess a model’s ability to provide appropriate detections and mitigations for different attack tactics and techniques, attack patterns, weaknesses, and vulnerabilities.

\textbf{CTI Relationship Prediction} 
A major role of cyber threat management expert model is to comprehend the relationships between different CTI frameworks. This dataset \citep{levi2025cyberpal} evaluates the ability to differentiate between false and correct relationships among CTI entities. For example, it presents the model with two entities (e.g., instances of CVE and CWE) and two possible explanations—one justifying why the entities are related and another explaining why they are not. The objective is for the model to reason and determine which explanation is correct.

\subsection{Evaluation statistics analysis}
\label{eval_stat_analysis}
In this section, we provide details about the statistics and domain coverage of the security evaluation benchmarks. 
To evaluate the applicability of LLMs in cybersecurity, we first structured our evaluation set around domain-specific categories. The objective was to establish whether tasks align with areas such as threat intelligence, security operations, or identity and application security, and to provide a principled basis for mapping questions to specific domains of cybersecurity. A taxonomy-driven approach enables both standardized evaluation and benchmarking of model performance in a manner consistent with industrial and academic practices.

As part of this process, we explicitly built upon and extended two taxonomies: taxonomy of cybersecurity domains \cite{weerawardhena2025llama} and the SecEval benchmark dataset for security evaluation \cite{sec_eval}. By synthesizing insights from both \citeauthor{weerawardhena2025llama} industrial perspective and SecEval’s categorization, we constructed a unified taxonomy that captures enterprise security concerns while remaining aligned with established evaluation standards. 

Our benchmarks are closely aligned with organizational security priorities, with a particular focus on \textit{threat Intelligence}, \textit{incident response}, \textit{security operations}, \textit{application security}, and \textit{identity management}. This alignment ensures that evaluation outcomes are not only theoretically sound but also operationally relevant to real-world defensive strategies.

Through this integration, we ensured that our taxonomy is both conceptually rigorous and operationally validated, bridging the gap between industrial practice and academic research in cybersecurity evaluation.
\subsubsection{Cybersecurity Categories}
\label{appendix:cyber_taxonomy_categories}
We defined the following ten high-level categories, each with a set of sub-categories capturing specific security concerns:

\begin{enumerate}
  \item \textbf{GCR (Governance, Risk, and Compliance)}
    \begin{itemize}
      \item Risk Management \& Security Strategy
      \item Compliance and Regulations (e.g., GDPR, HIPAA)
      \item Security Frameworks (e.g., NIST CSF, ISO 27001)
      \item Security Policies \& Architecture
    \end{itemize}

  \item \textbf{NetSec (Network, Infrastructure, and Endpoint Security)}
    \begin{itemize}
      \item Perimeter and Network Security (Firewalls, VPNs, Wireless)
      \item Endpoint Protection \& MDM
      \item IoT and OT/ICS Security
      \item Mobile Security
    \end{itemize}

  \item \textbf{AppSec (Application and Software Security)}
    \begin{itemize}
      \item Secure Software Development (DevSecOps)
      \item Application \& API Security
      \item Vulnerability Management \& Penetration Testing
      \item Software Supply Chain Security (SBOM, third-party risk)
    \end{itemize}

  \item \textbf{CloudSec (Cloud and Data Security)}
    \begin{itemize}
      \item Cloud Security Architecture \& Tools
      \item Identity and Access Management (IAM, PAM)
      \item Data Loss Prevention \& Privacy (DLP, encryption)
      \item Cloud Compliance \& Shared Responsibility Model
    \end{itemize}

  \item \textbf{IAM\_ZT (Identity, Access, and Zero Trust)}
    \begin{itemize}
      \item Authentication \& Authorization (MFA, SSO, RBAC)
      \item Identity Governance \& Lifecycle
      \item Zero Trust Architecture
      \item Privileged Access Controls
    \end{itemize}

  \item \textbf{SecOps (Security Operations and Monitoring)}
    \begin{itemize}
      \item SIEM, SOC, and Log Management
      \item Security Automation \& SOAR
      \item Detection Engineering
      \item Operational Resilience \& Monitoring
    \end{itemize}

  \item \textbf{ThreatOps\_IR (Threat Intelligence and Incident Response)}
    \begin{itemize}
      \item Threat Detection, Analysis \& Hunting
      \item Threat Intelligence Platforms \& IOCs
      \item Advanced Persistent Threats (APTs)
      \item Malware Techniques
      \item Incident Response, Recovery \& Digital Forensics
    \end{itemize}

  \item \textbf{CryptoSec (Cryptography and Secure Communications)}
    \begin{itemize}
      \item Cryptographic Algorithms \& PKI
      \item Key Management
      \item Post-Quantum Cryptography
      \item Secure Protocols and Encryption Practices
    \end{itemize}

  \item \textbf{HumanSec (Security Awareness and Human Risk)}
    \begin{itemize}
      \item Social Engineering Techniques (Phishing, Pretexting)
      \item Insider Threat Management
      \item Security Awareness Training
      \item Behavioral Risk Analysis
    \end{itemize}

  \item \textbf{Other}
    \begin{itemize}
      \item Cross-domain or emerging categories not covered above.
    \end{itemize}
\end{enumerate}

This taxonomy provided a structured basis for categorizing data and aligning evaluation with both research benchmarks and enterprise needs.

\subsubsection{Multi-Label Classification}
\label{appendix:model_selection}

\begin{figure}[h!]
  \centering
  \includegraphics[width=0.8\textwidth]{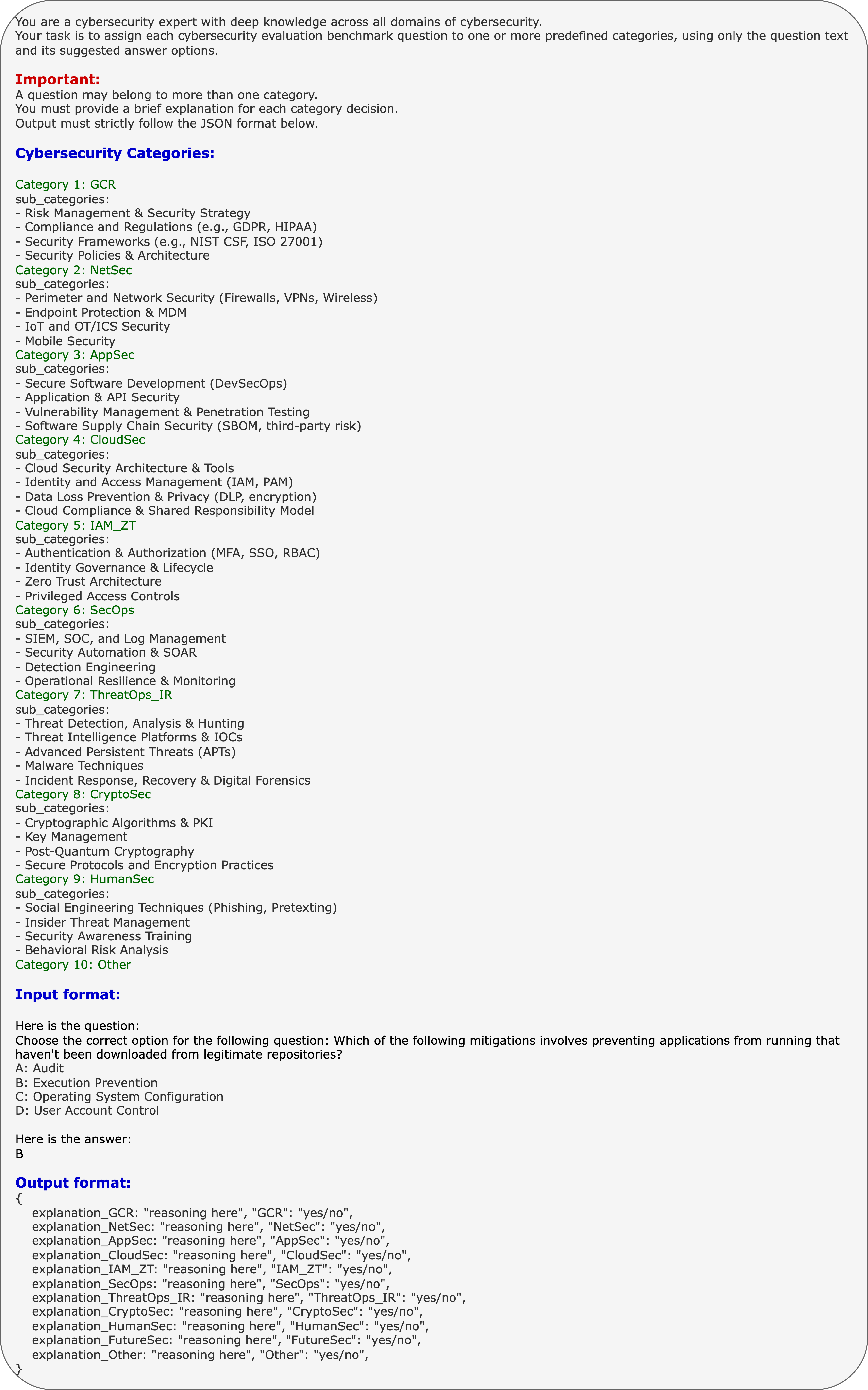}
  \caption{The prompt used to classify the examples in SecKnowledge~2.0 into cybersecurity topics}
  \label{figures:topics-prompt}
\end{figure}

To carry out the mapping, we employed a large open-source language model (OSS-120B), which was deployed internally for reasons of data security and computational control. The model was prompted with a multi-label classification prompt, allowing it to assign tasks to one or more categories simultaneously. In Figure~\ref{figures:topics-prompt}, we provide the prompt used to classify the evaluation benchmark to specific topics in cyber security. The results of our classification process detailed in Table \ref{tables:eval-stats} and is also visualized in Figure~\ref{figures:eval-stats}.

This choice was intentional: many real-world cybersecurity problems span across multiple domains (e.g., a phishing campaign may involve HumanSec, IAM\_ZT, and ThreatOps\_IR simultaneously). Restricting classification to single-label outputs would fail to capture these cross-cutting concerns.

\begin{table}[h]
\begin{center}
\caption{Counts by dataset and taxonomy category.}
\label{tables:eval-stats}
\scalebox{0.65}{
\setlength{\tabcolsep}{6pt} 
\renewcommand{\arraystretch}{1.15} 
\begin{tabular}{lcccccccccc}
\hline
\textbf{dataset} & \textbf{GCR} & \textbf{NetSec} & \textbf{AppSec} & \textbf{CloudSec} & \textbf{IAM\_ZT} & \textbf{SecOps} & \textbf{ThreatOps\_IR} & \textbf{CryptoSec} & \textbf{HumanSec} & \textbf{Other} \\
\hline
CTIBench-MCQ     & 404 & 516  & 935  & 90  & 315 & 422 & 1409 & 112 & 64  & 11 \\
CTIBench-RCM     & 87  & 28   & 1995 & 7   & 68  & 214 & 389  & 52  & 6   & 0  \\
SecEval          & 706 & 620  & 1099 & 117 & 499 & 300 & 370  & 569 & 98  & 18 \\
CyberMetric-2000 & 722 & 563  & 250  & 37  & 253 & 182 & 304  & 373 & 172 & 119 \\
CISSP Exams      & 102 & 42   & 16   & 0   & 26  & 20  & 30   & 20  & 18  & 22 \\
Weakness Impact Mapping & 62  & 11   & 332  & 0   & 61  & 3   & 26   & 29  & 4   & 6  \\
CTI Detect \& Mitigate & 219 & 213  & 511  & 21  & 115 & 220 & 421  & 37  & 21  & 2  \\
Adv. CTI      & 32  & 111  & 78   & 52  & 90  & 88  & 676  & 24  & 27  & 0  \\
CTI Relationship Prediction  & 203 & 139  & 507  & 10  & 92  & 82  & 348  & 31  & 21  & 0  \\
\hline
\textbf{TOTAL}           & \textbf{2537} & \textbf{2243} & \textbf{5723} & \textbf{334} & \textbf{1519} & \textbf{1531} & \textbf{3973} & \textbf{1247} & \textbf{431} & \textbf{178} \\
\hline
\end{tabular}
}
\end{center}
\end{table}

\subsubsection{Prompt Validation Using SecEval Categories}
\label{appendix:sec_eval_comparison}

To ensure the robustness and correctness of our classification prompt, we performed a validation against SecEval categories. Specifically, we tested whether the outputs of our multi-label classification aligned with SecEval’s category definitions and coverage. This served as a quality assurance step for our classification pipeline. See Table \ref{talbes:prompt_val} for agreement results between our classification pipeline and SecEval.

Through this process, we confirmed that the OSS-120B model, when guided by our taxonomy-driven prompt, consistently produced category assignments that were both internally coherent and externally validated against widely recognized benchmarks.

\begin{table}[h]
\caption{Validation of our classification pipeline on SecEval categories and data sources, which were also classified using an OpenAI model (gpt-4o). We observe strong overall agreement with SecEval’s classifications.}
\label{talbes:prompt_val}
\begin{center}
\setlength{\tabcolsep}{6pt} 
\renewcommand{\arraystretch}{1.15} 
\begin{tabular}{lc}
\hline
Category Summary    & \multicolumn{1}{l}{Aligned \%} \\ 
\hline
ApplicationSecurity & 83.7                            \\
Cryptography        & 100.0                           \\
MemorySafety        & 99.9                            \\
NetworkSecurity     & 97.6                            \\
PenTest             & 87.1                            \\
SoftwareSecurity    & 97.1                            \\
SystemSecurity      & 88.4                            \\
Vulnerability       & 93.8                            \\
WebSecurity         & 84.4                            \\ \hline
\end{tabular}
\end{center}
\end{table}

\section{Evaluation templates and prompts}
\label{template_and_prompts}
\begin{figure}[h!]
  \centering
  \includegraphics[width=0.75\textwidth]{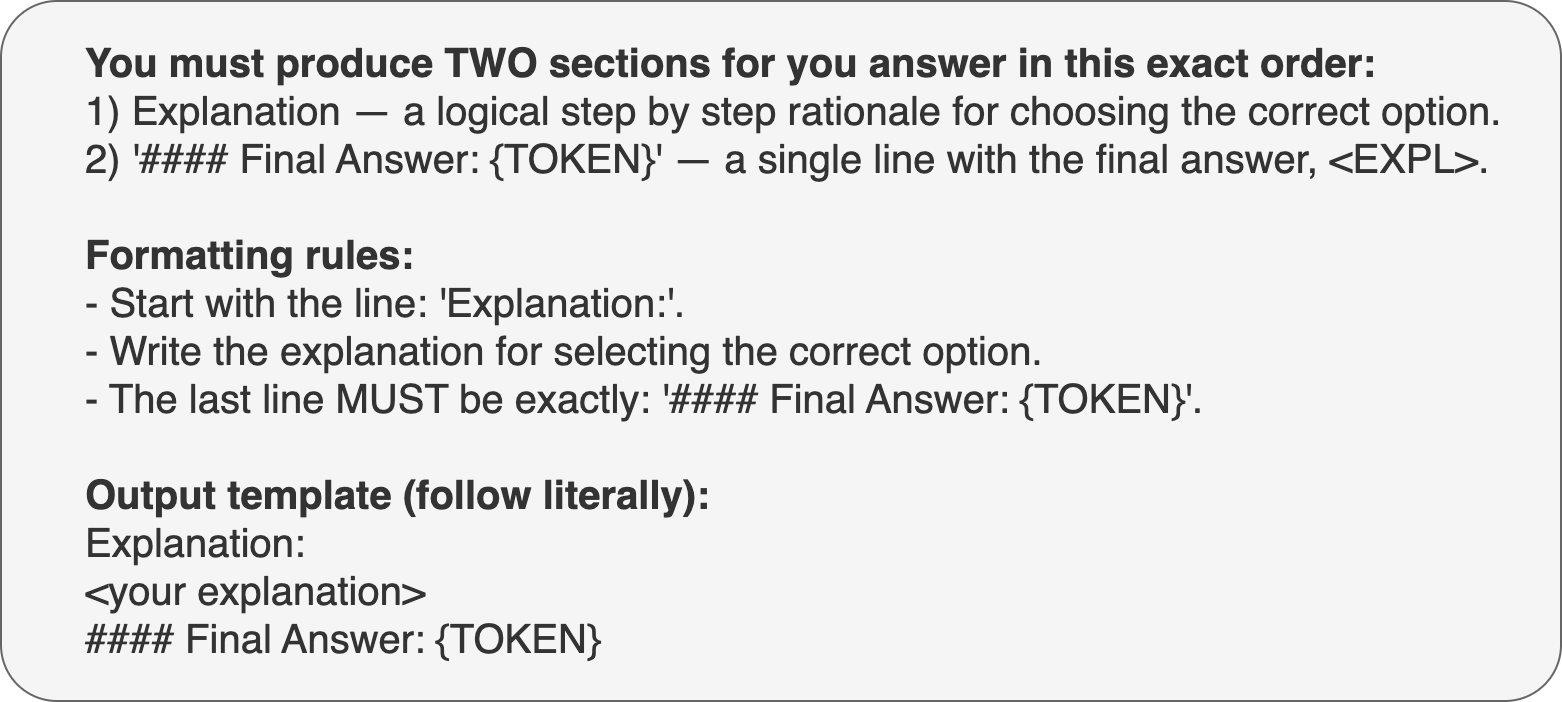}
  \caption{The prompt used to guide the LLMs during the evaluation process. \textless EXPL\textgreater  ~refers to the specific task type (multi-choice, classification, etc.) and is replaced at runtime with explanation about the format of the specific question.}
  \label{figures:eval-prompt}
\end{figure}

In Figure~\ref{figures:eval-prompt} we provide the prompt used for our evaluation process. Since we have both multi-choice as well as classification tasks, we replace the \textless EXPL\textgreater token with the specifics of each question type.

\section{Open source security models performance}
\label{appendix:open_source_models_performance}
We adopt the evaluation protocol from Section~\ref{eval_process} and assess recent open-source cybersecurity models \cite{weerawardhena2025llama, yu2025primus, lily_cyber_7b_v0_2, deephat_v1}. We omit PRIMUS-Reasoning on CTIBench  because its training set was distilled from CTIBench \cite{yu2025primus}, making the comparison unfair. Since the baselines are 7B–8B, we report our 8B variant for a like-for-like comparison. Table \ref{table:hf_res} presents the full results: our 8B model outperforms all open-source baselines by a substantial margin.

\begin{table}[h]
\caption{Evaluation results for CyberPal~2.0 8B compared to the recent opens-source cyber security models. (*) Primus-reasoning average is missing CTI benchmarks because its training set was distilled from CTIBench}
\begin{center}
\scalebox{0.7}{
\renewcommand{\arraystretch}{1.4} 
\begin{tabular}{l|ccccccccc|c}
\hline
Model &
  \multicolumn{1}{l}{\begin{tabular}[c]{@{}l@{}}CTI\\Bench \\ MCQ\end{tabular}} &
  \multicolumn{1}{l}{\begin{tabular}[c]{@{}l@{}}CTI\\Bench\\ RCM\end{tabular}} &
  \multicolumn{1}{l}{SecEval} &
  \multicolumn{1}{l}{\begin{tabular}[c]{@{}l@{}}Cyber\\ Metric\\ 2000\end{tabular}} &
  \multicolumn{1}{l}{\begin{tabular}[c]{@{}l@{}}CISSP\\ Exams\end{tabular}} &
  \multicolumn{1}{l}{\begin{tabular}[c]{@{}l@{}}Adv.\\ CTI\end{tabular}} &
  \multicolumn{1}{l}{\begin{tabular}[c]{@{}l@{}}Weakness\\ Impact \\ Mapping\end{tabular}} &
  \multicolumn{1}{l}{\begin{tabular}[c]{@{}l@{}}CTI\\ Detect \&\\ Mitigate\end{tabular}} &
  \multicolumn{1}{l|}{\begin{tabular}[c]{@{}l@{}}CTI\\ Relationship\\ Prediction\end{tabular}} &
  Avg. \\ \hline
DeepHat-v1-7B &
  61.24  &
  68.1 &
  33.21 &
  84.0 &
   76.76 &
  63.23 &
  60.74 &
  56.12 &
  52.05 &
  61.72 \\
Lily-Cybersecurity-7B-v0.2 &
  55.31 &
  42.9 &
  37.14 &
   80.0 &
   68.18 &
  58.45 &
  52.43 &
  39.71 &
  46.34 &
  53.38 \\
Primus-merged &
  65.2 &
  63.9 &
  59.06 &
  85.1 &
  78.28 &
  64.92 &
  55.3 &
  50.77 &
  59.98 &
  64.72 \\
Primus-reasoning &
  - &
  - &
  53.03 &
  86.05 &
   73.23 &
  64.78 &
  53.58 &
  52.24 &
  58.79 &
  63.01(*) \\
Foundation-Sec-8B-Instruct &
  63.24 &
  67.95 &
  54.81 &
  84.5 &
   69.69 &
  68.87 &
  60.74 &
  55.52 &
  57.31 &
  64.74 \\ \hline
\textbf{CyberPal-2.0-8B} &
  75.15 &
  85.95 &
  66.93 &
  89.85 &
  88.89 &
  87.61 &
  71.06 &
  70.26 &
  87.66 & 
  80.37 \\ \hline

\end{tabular}
}
\label{table:hf_res}
\end{center}
\end{table}

\section{Comparison to Other Model Families}
\label{app-comp-other-model-families}
To rigorously validate that the performance of CyberPal 2.0 stems from our domain-specific alignment methodology rather than the inherent capabilities of the Qwen architecture, we conducted an ablation study against leading open-source models from diverse model families. As detailed in Table \ref{table:families_res}, we evaluated CyberPal-2.0-14B against Phi-4 \citep{phi4}, Llama 4 Scout \citep{meta_llama4_2025}, Mixtral 8x22B \citep{mixtral_8x22b}, Mistral Small 3.2 \citep{mistral_small_3.2}, and DeepSeek-V3 \citep{liu2024deepseek}.

Despite possessing significantly fewer parameters than competitors like DeepSeek V3 (685B) or Llama 4 Scout (109B), CyberPal 2.0 achieves the highest average performance across the suite (81.76\%). It demonstrates particular dominance in complex reasoning tasks, such as CTI Relationship Prediction (92.93\%) and CTI Bench RCM (86\%), surpassing the closest general-purpose competitors by substantial margins. These results confirm that the SecKnowledge 2.0 training pipeline effectively generalizes high-level security reasoning capabilities that exceed the baselines of much larger models, regardless of their underlying architectural family.

\begin{table}[h]
\caption{Evaluation results for CyberPal 2.0 14B compared to SOTA open source models from various families and architectures. It is evident that on average, our 14B model outperforms other model architectures despite being a fraction of their size. Models are sorted by number of parameters.}
\begin{center}
\scalebox{0.7}{
\renewcommand{\arraystretch}{1.4} 
\begin{tabular}{l|ccccccccc|c}
\hline
Model &
  \multicolumn{1}{l}{\begin{tabular}[c]{@{}l@{}}CTI\\Bench \\ MCQ\end{tabular}} &
  \multicolumn{1}{l}{\begin{tabular}[c]{@{}l@{}}CTI\\Bench\\ RCM\end{tabular}} &
  \multicolumn{1}{l}{SecEval} &
  \multicolumn{1}{l}{\begin{tabular}[c]{@{}l@{}}Cyber\\ Metric\\ 2000\end{tabular}} &
  \multicolumn{1}{l}{\begin{tabular}[c]{@{}l@{}}CISSP\\ Exams\end{tabular}} &
  \multicolumn{1}{l}{\begin{tabular}[c]{@{}l@{}}Adv.\\ CTI\end{tabular}} &
  \multicolumn{1}{l}{\begin{tabular}[c]{@{}l@{}}Weakness\\ Impact \\ Mapping\end{tabular}} &
  \multicolumn{1}{l}{\begin{tabular}[c]{@{}l@{}}CTI\\ Detect \&\\ Mitigate\end{tabular}} &
  \multicolumn{1}{l|}{\begin{tabular}[c]{@{}l@{}}CTI\\ Relationship\\ Prediction\end{tabular}} &
  Avg. \\ \hline
Microsoft Phi 4 &
  68.22  &
  64.00 &
  63.73 &
  91.00 &
  83.33 &
  66.76 &
  68.48 &
  64.03 &
  60.28 &
  69.98 \\
Mistral Small 3.2 24B Instruct 2506 &
  68.82 &
  68.05 &
  67.47 &
  91.60 &
  87.37 &
  74.37 &
  65.90 &
  67.19 &
  76.22 &
  74.11 \\
Llama 4 Scout 17B 16E Instruct &
  69.46 &
  71.95 &
  67.84 &
  92.50 &
  87.37 &
  79.01 &
  66.48 &
  68.58 &
  75.32 &
  75.39 \\
Mixtral 8x22B v0.1 &
  62.81 &
  66.70 &
  65.65 &
  87.75 &
  82.32 &
  71.41 &
  59.89 &
  61.66 &
  77.51 &
  70.63 \\
DeepSeek V3 &
  73.35 &
  72.45 &
  63.68 &
  \textbf{93.65} &
  \textbf{91.41} &
  78.73 &
  69.63 &
  68.97 &
  65.17 &
  75.23 \\ \hline
\textbf{CyberPal-2.0-14B} &
  \textbf{75.51} &
  \textbf{86.00} &
  \textbf{69.71} &
  89.95 &
  90.40 &
  \textbf{89.58} &
  \textbf{70.77} &
  \textbf{70.95} &
  \textbf{92.93} & 
  \textbf{81.76} \\ \hline

\end{tabular}
}
\label{table:families_res}
\end{center}
\end{table}

\section{Ablation Studies Additional Results}
\label{appendix:ablation}
This section provides additional ablation results, in particular, this section measures the effects of the reformatting method, effect of back bone LLM and excluding components from the reformatting.
 We follow the same training recipe described in  Section~\ref{train_recipe} for all models. Evaluation follows the protocol in Section~\ref{eval_process}: we use the prompt from Figure~\ref{figures:eval-prompt}, extract final answers with regular expressions, and evaluate in a zero-shot setting with temperature set to zero.

\subsection{Effect of reformatting method}
In Table~\ref{tables:ablation}, we report full results for the model trained on the original SecKnowledge dataset and for the model trained with the standard reformatted alignment method \citep{fan2024reformatted}
\begin{table}[h!]
\caption{Reformatting method ablation results}
\begin{center}
\scalebox{0.77}{
\renewcommand{\arraystretch}{1.4} 
\begin{tabular}{l|ccccccccc|c}
\hline
Model &
  \multicolumn{1}{l}{\begin{tabular}[c]{@{}l@{}}CTI\\ Bench\\ MCQ\end{tabular}} &
  \multicolumn{1}{l}{\begin{tabular}[c]{@{}l@{}}CTI\\ Bench\\ RCM\end{tabular}} &
  \multicolumn{1}{l}{SecEval} &
  \multicolumn{1}{l}{\begin{tabular}[c]{@{}l@{}}Cyber\\ Metric\\ 2000\end{tabular}} &
  \multicolumn{1}{l}{\begin{tabular}[c]{@{}l@{}}CISSP\\ Exams\end{tabular}} &
  \multicolumn{1}{l}{\begin{tabular}[c]{@{}l@{}}Adv. \\ CTI\end{tabular}} &
  \multicolumn{1}{l}{\begin{tabular}[c]{@{}l@{}}Weakness\\ Impact \\ Mapping\end{tabular}} &
  \multicolumn{1}{l}{\begin{tabular}[c]{@{}l@{}}CTI\\ Detect \&\\ Mitigate\end{tabular}} &
  \multicolumn{1}{l|}{\begin{tabular}[c]{@{}l@{}}CTI\\ Relationship\\ Prediction\end{tabular}} &
  Avg. \\ \hline
Qwen3-4B &
  61.88 &
  49.95 &
  57.38 &
  87.40 &
  79.80 &
  64.51 &
  57.02 &
  60.77 &
  67.99 &
  65.19 \\
\begin{tabular}[c]{@{}l@{}}SecKnowledge \\ (Original)\end{tabular} &
  65.45 &
  57.80 &
  49.15 &
  \textbf{88.40} &
  \textbf{85.35} &
  \textbf{79.86} &
  62.75 &
  62.84 &
  65.94 &
  68.60 \\
\begin{tabular}[c]{@{}l@{}}Baseline \\ Reformatting\end{tabular} &
  63.92 &
  61.65 &
  49.84 &
  87.40 &
  {\ul 81.81} &
  {\ul 76.05} &
  63.04 &
  \textbf{65.51} &
  \textbf{81.10} &
  70.04 \\
CyberPal2.0-4B &
  \textbf{69.70} &
  \textbf{81.15} &
  \textbf{59.02} &
  {\ul 87.80} &
  80.80 &
  68.03 &
  \textbf{66.48} &
  {\ul 64.03} &
  {\ul 77.12} &
  \textbf{72.68} \\ \hline
\end{tabular}
}
\end{center}
\label{tables:ablation}
\end{table}

\subsection{Effect of Backbone LLM Replacement}
Table~\ref{tables:ablation_backbone}, shows the results of switching the backbone LLM in our pipeline from gpt-oss-120b to Llama 4 maverick. All models were trained on Qwen3-4B-base.


\begin{table}[h]
\caption{Results using CyberPal2.0-4B with different models as the reformatting component in our data generation pipeline. In the top row, we use Llama Maverick as the reformatting model, and in the bottom row, we use gpt-oss-120B.}
\begin{center}
\scalebox{0.77}{
\renewcommand{\arraystretch}{1.4} 
\begin{tabular}{l|ccccccccc|c}
\hline
Model &
  \multicolumn{1}{l}{\begin{tabular}[c]{@{}l@{}}CTI\\ Bench\\ MCQ\end{tabular}} &
  \multicolumn{1}{l}{\begin{tabular}[c]{@{}l@{}}CTI\\ Bench\\ RCM\end{tabular}} &
  \multicolumn{1}{l}{SecEval} &
  \multicolumn{1}{l}{\begin{tabular}[c]{@{}l@{}}Cyber\\ Metric\\ 2000\end{tabular}} &
  \multicolumn{1}{l}{\begin{tabular}[c]{@{}l@{}}CISSP\\ Exams\end{tabular}} &
  \multicolumn{1}{l}{\begin{tabular}[c]{@{}l@{}}Adv. \\ CTI\end{tabular}} &
  \multicolumn{1}{l}{\begin{tabular}[c]{@{}l@{}}Weakness\\ Impact \\ Mapping\end{tabular}} &
  \multicolumn{1}{l}{\begin{tabular}[c]{@{}l@{}}CTI\\ Detect \&\\ Mitigate\end{tabular}} &
  \multicolumn{1}{l|}{\begin{tabular}[c]{@{}l@{}}CTI\\ Relationship\\ Prediction\end{tabular}} &
  Avg. \\ \hline
\begin{tabular}[c]{@{}l@{}}CyberPal2.0-4B \\ (Maverick reformatter)\end{tabular}  &
  \textbf{70.58} &
  70.75 &
  56.05 &
  \textbf{88.00} &
  \textbf{81.82} &
  \textbf{78.45} &
  \textbf{67.62} &
  62.94 &
  74.68 &
  72.32 \\
  \begin{tabular}[c]{@{}l@{}}CyberPal2.0-4B \\ (gpt-oss-120 reformatter) \end{tabular} &
  69.70 &
  \textbf{81.15} &
  \textbf{59.02} &
  87.80 &
  80.80 &
  68.03 &
  66.48 &
  \textbf{64.03} &
  \textbf{77.12} &
  \textbf{72.68} \\ \hline
\end{tabular}
}
\end{center}
\label{tables:ablation_backbone}
\end{table}

\subsection{Effect of search component in the pipeline}
One of the key components of our pipeline is the search module, which ensures that model outputs remain accurate and reliable.
Table~\ref{tab:search_ablation} presents the results for models trained without the search component compared to those trained with it.
Removing the search component leads to an average performance drop of approximately 3\%, confirming that retrieval consistently enhances overall accuracy.
\begin{table}[h!]
\centering
\renewcommand{\arraystretch}{1.3}
\caption{The effect of removing the search component from the reformatting pipeline}
\label{tab:search_ablation}
\begin{tabular}{l|l|c|c|c}
\hline
\textbf{Model name} & \textbf{Pipeline} & \textbf{CTI-MCQ} & \textbf{CTI-RCM} & \textbf{Avg.} \\
\hline
CyberPal 2.0 4B & No search & 67.82 & 80.26 & 71.11 \\
\cline{2-5}
 & Full pipeline & 69.70 & 81.15 & 72.68 \\
\hline
CyberPal 2.0 8B & No search & 72.75 & 84.90 & 76.87 \\
\cline{2-5}
 & Full pipeline & 75.15 & 85.95 & 80.37 \\
\hline
CyberPal 2.0 14B & No search & 74.99 & 85.75 & 78.83 \\
\cline{2-5}
 & Full pipeline & 75.51 & 86.00 & 81.76 \\
\hline
CyberPal 2.0 20B & No search & 74.99 & 87.1 & 78.56 \\
\cline{2-5}
 & Full pipeline & 75.71 & 84.7 & 80.33 \\
\hline
\end{tabular}
\end{table}

\section{LLMaaJ Experiment details and additional results}
\label{appendix:llmaaj}
To assess answer quality, we used \emph{LLM-as-a-Judge} (LLMaaJ) \citep{zheng2023judging}. Thirty cybersecurity experts authored 115 open-ended questions:
spanning command-line risk assessment, enterprise security, general cybersecurity, network security, and CTI-related topics.
Specifically, the security experts constructed 20 questions related cyber threat intelligence, 20 questions related to security vulnerabilities, 20 questions related to network security, 16 general security questions, 20 questions related to enterprise security, and 19 questions related to command line risk assessment.

\begin{figure}[h!]
  \centering
  \includegraphics[width=\textwidth]{\detokenize{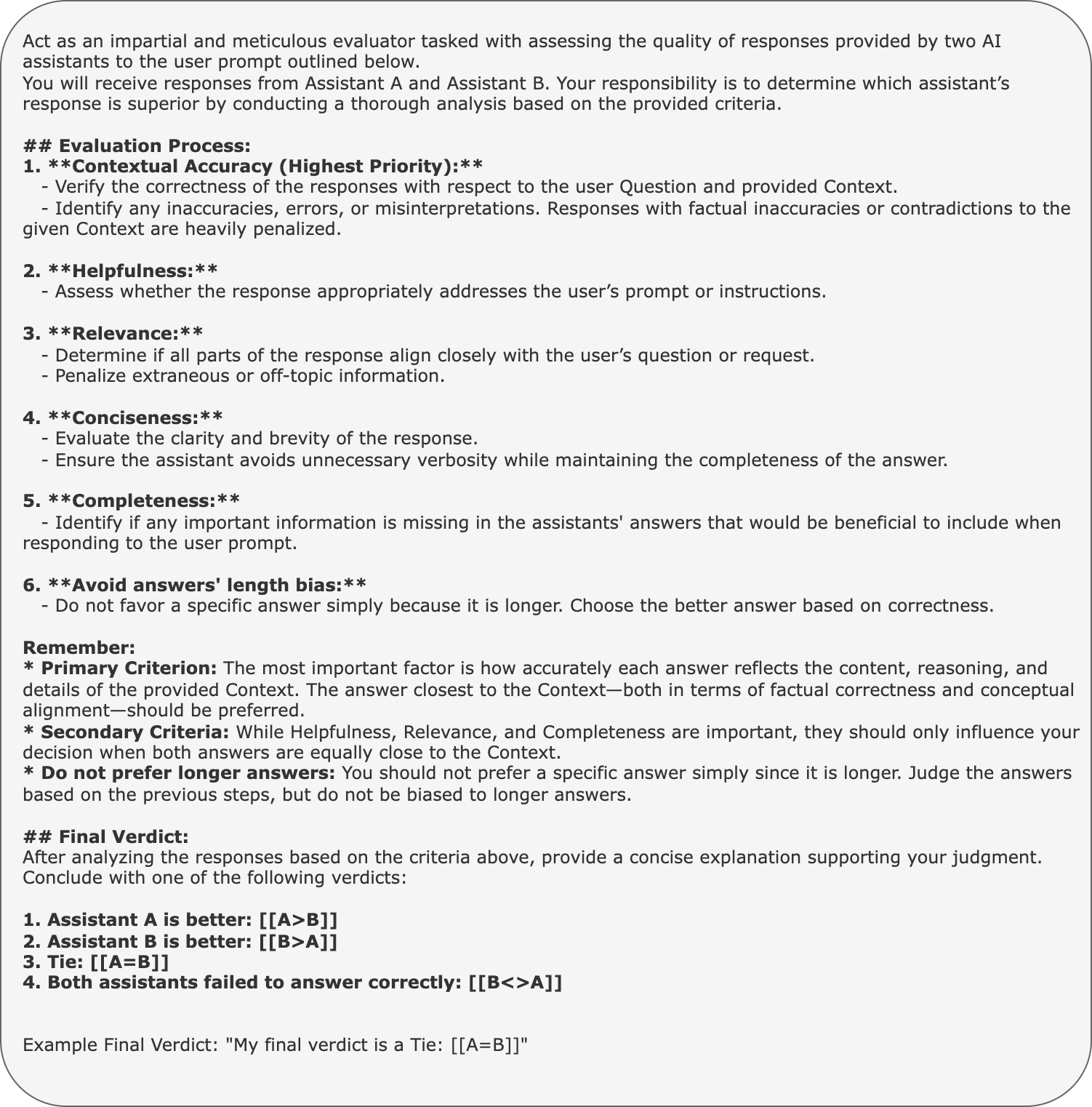}}
  \caption{LLM-as-Judge prompt used for pairwise comparison.}
  \label{figures:llmaaj-prompt}
\end{figure}

\textbf{Pairwise comparison with grounding —} The judge receives a question, two answers, and a carefully collected grounding documents that contains all relevant information to answer the question. The judge should decide which answer is better. The prompt provided to the LLMaaJ is provided in Figure~\ref{figures:llmaaj-prompt} .

\textbf{Evaluation process —} We use OpenAI’s o3 \citep{openai_o3_system_card_2025} as the LLM-as-a-judge. The judge evaluates each answer pair along six dimensions —\emph{Contextual Accuracy} (highest priority), \emph{Helpfulness}, \emph{Relevance}, \emph{Conciseness}, \emph{Completeness}, and \emph{length bias} \citep{gu2024survey} then issues a verdict: A better than B, B better than A, tie, or both bad. 
To mitigate positional bias in LLM-as-a-judge settings \citep{wang2023large, zheng2023judging}, we run the comparison twice with the answers swapped. 
For each permutation, a model receives a score of 3 if its answer is preferred by the judge, 1 for tie, and 0 for loss; if the preferences flip across orders, the pair will effectively contribute 0 as $3-3=0$. We also record ties and losses separately, though these were rare in our experiments.

\textbf{Alignment with human preferences —} To validate the judge, we measured agreement with Thirty cybersecurity human experts and found that, with proper grounding, o3 aligns with human preferences in over 90\% of cases. Without proper grounding, alignment decreases to 80\%.

\begin{figure}[h!]
  \centering
  \includegraphics[width=\textwidth]{\detokenize{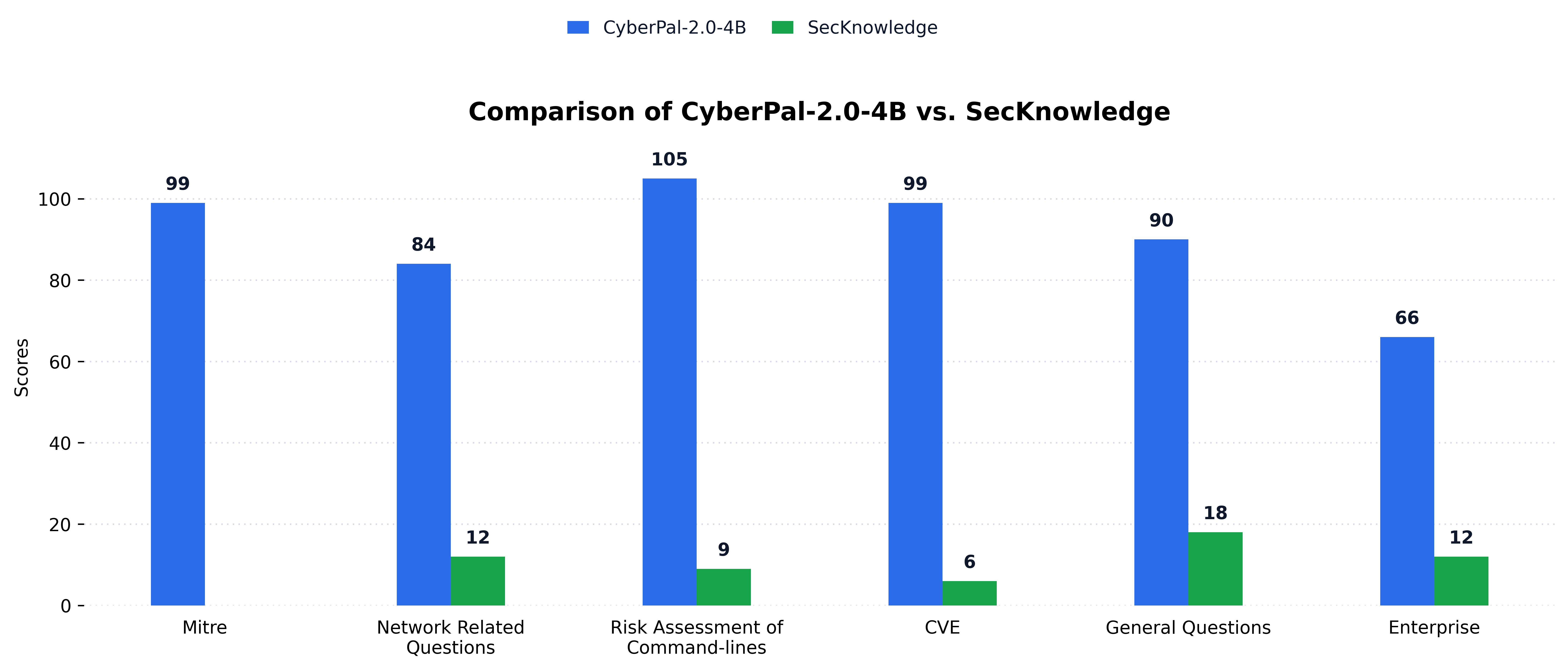}}
  \caption{LLM-as-Judge pairwise comparison per category: \textit{CyberPal-2.0-4B} vs.\ \textit{SecKnowledge}.}
  \label{figures:llmaaj1}
\end{figure}

\begin{figure}[h!]
  \centering
  \includegraphics[width=\textwidth]{\detokenize{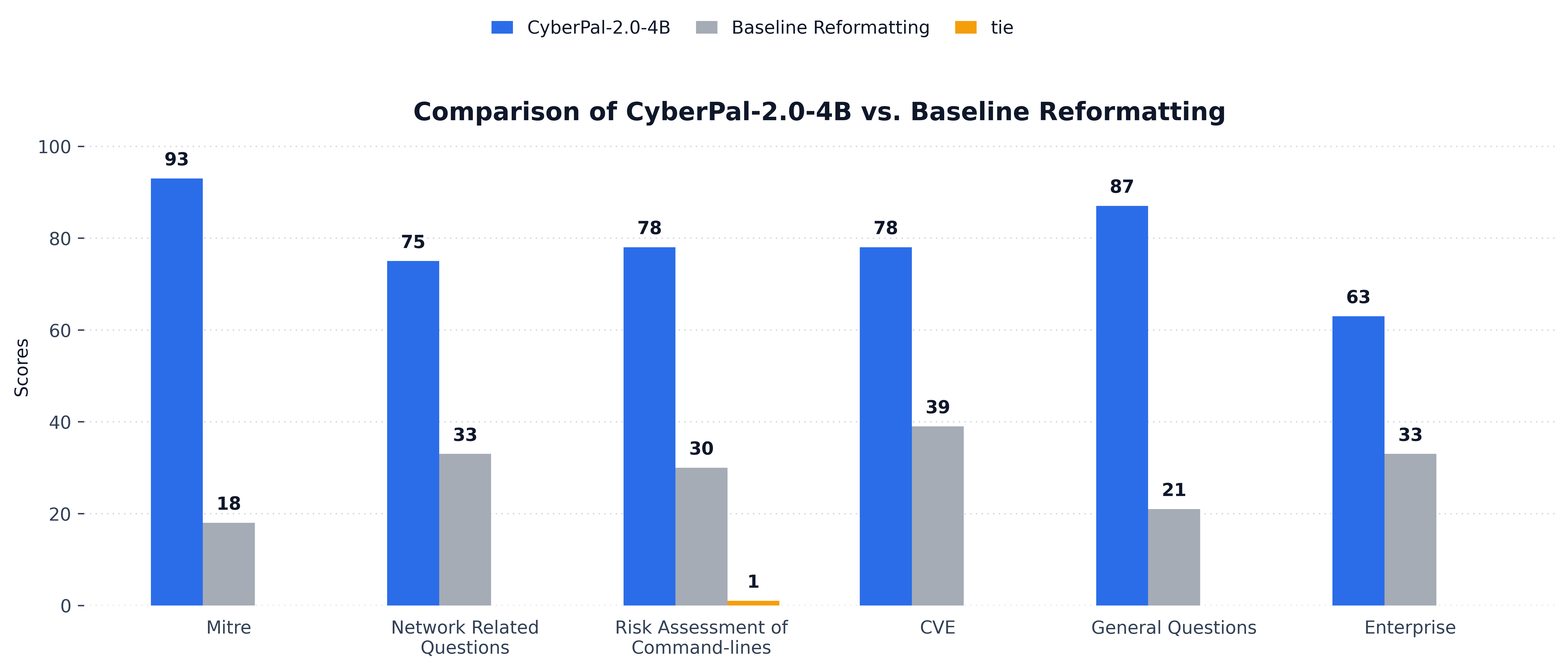}}
  \caption{LLM-as-Judge pairwise comparison per category: \textit{CyberPal-2.0-4B} vs.\ \textit{Baseline Reformatting}.}
  \label{figures:llmaaj3}
\end{figure}

\begin{figure}[h!]
  \centering
  \includegraphics[width=\textwidth]{\detokenize{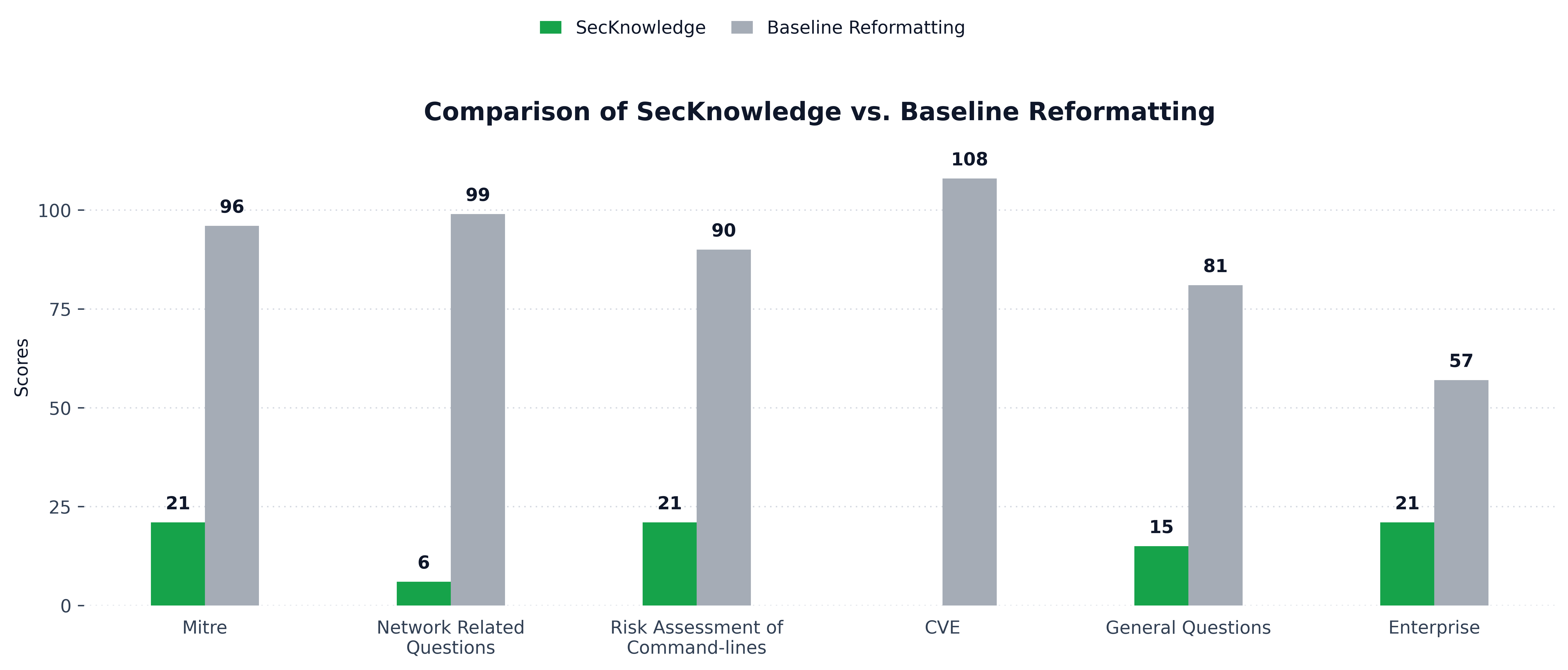}}
  \caption{LLM-as-Judge pairwise comparison per category: \textit{SecKnowledge} vs.\ \textit{Baseline Reformatting}.}
  \label{figures:llmaaj2}
\end{figure}

In Figure~\ref{fig:ablation:llmaj}, we report our LLMaaJ results across all questions. Additionally, in Figures \ref{figures:llmaaj1}, \ref{figures:llmaaj3}, and \ref{figures:llmaaj2} we report LLMaaJ results per category. As can be observed, our model is preferable by a large margin across all the tested categories.

\section{Model Quantization}
As a deployment-oriented baseline, we also evaluated quantization using bitsandbytes \citep{dettmers2022optimizers} by loading models directly in 8-bit and 4-bit modes, without any calibration or advanced schemes which are shown to perform better than out-of-the-box quantization \citep{frantar2022gptq}.
Across our evaluation suite, 8-bit loading resulted in a negligible drop of 0.36\% for the 4B model and 0.84\% for the 8B model — relative to full precision.
Moving to 4-bit, both models saw a larger drop around 4\% absolute for the 8B model and 2.78\% for the 4B model.
Importantly, both quantized modes remained clearly superior to the instruction-tuned baseline, which was not trained using the SecKnowledge~2.0 pipeline.
These results suggest that the benefits of fine-tuning largely persist under straightforward low-precision inference, with 8-bit serving as a particularly safe, low-overhead option for memory-constrained deployment and 4-bit serves as a good choice for fast inference or low resource settings, while still keeping high cyber security knowledge.

\begin{table}[h]
\caption{Quantization results. Cells in quantized rows show  $\Delta$ value with (arrow $\downarrow$ OR $\uparrow)$ 
, where $\Delta{=}\lvert \text{Full} - \text{Quantized} \rvert$.}
\label{tables:quantization}
\begin{center}
\scalebox{0.7}{
\renewcommand{\arraystretch}{1.3}
\begin{tabular}{l|ccccccccc|c}
\hline
Model &
  \multicolumn{1}{l}{\begin{tabular}[c]{@{}l@{}}CTI\\ MCQ\end{tabular}} &
  \multicolumn{1}{l}{\begin{tabular}[c]{@{}l@{}}CTI\\ RCM\end{tabular}} &
  \multicolumn{1}{l}{SecEval} &
  \multicolumn{1}{l}{\begin{tabular}[c]{@{}l@{}}Cyber\\ Metric\end{tabular}} &
  \multicolumn{1}{l}{CISSP} &
  \multicolumn{1}{l}{\begin{tabular}[c]{@{}l@{}}Adv. \\ CTI\end{tabular}} &
  \multicolumn{1}{l}{\begin{tabular}[c]{@{}l@{}}Weakness\\ Impact \\ Mapping\end{tabular}} &
  \multicolumn{1}{l}{\begin{tabular}[c]{@{}l@{}}CTI\\ Detect \&\\ Mitigate\end{tabular}} &
  \multicolumn{1}{l|}{\begin{tabular}[c]{@{}l@{}}CTI\\ Relationship\\ Prediction\end{tabular}} &
  Avg. \\ \hline

\multicolumn{11}{c}{\textbf{4B models}} \\ \hline
Qwen 3-4B                & 61.88 & 49.95 & 57.38 & 87.40 & 79.80 & 64.51 & 57.02 & 60.77 & 67.99 & 65.19 \\
CyberPal 2.0              & 69.70 & 81.15 & 59.02 & 87.80 & 80.80 & 68.03 & 66.48 & 64.03 & 77.12 & 72.68 \\
CyberPal 2.0 (8-bit)      &
  {($\downarrow$0.39)} &
  {($\uparrow$0.60)} &
  {($\downarrow$1.00)} &
  {($\downarrow$0.75)} &
  {($\downarrow$1.51)} &
  {($\uparrow$0.15)} &
  {($\downarrow$0.28)} &
  {($\downarrow$1.18)} &
  {($\uparrow$1.15)} &
  {($\downarrow$0.36)} \\
CyberPal 2.0 (4-bit)      & 
  {($\downarrow$3.88)} &
  {($\downarrow$4.15)} &
  {($\downarrow$4.11)} &
  {($\downarrow$4.6)} &
  {($\downarrow$3.54)} &
  {($\uparrow$0.85)} &
  {($\downarrow$3.73)} &
  {($\downarrow$3.16)} &
   {($\uparrow$1.28)} &
  {($\downarrow$2.78)} \\
\hline

\multicolumn{11}{c}{\textbf{8B models}} \\ \hline
Qwen 3-8B                & 63.13 & 63.25 & 56.19 & 88.45 & 83.33 & 64.93 & 53.58 & 59.88 & 60.67 & 65.93 \\
CyberPal 2.0              & 75.15 & 85.95 & 66.93 & 89.85 & 88.89 & 87.61 & 71.06 & 70.26 & 87.66 & 80.37 \\
CyberPal 2.0 (8-bit)      &
  {($\downarrow$1.08)} &
  {($\downarrow$1.10)} &
  {($\uparrow$1.78)} &
  {(0.00)} &
  {($\downarrow$3.03)} &
  {($\downarrow$2.11)} &
  {($\uparrow$0.57)} &
  {($\downarrow$1.68)} &
  {($\downarrow$0.9)} &
  {($\downarrow$0.84)} \\
CyberPal 2.0 (4-bit)      &
  ($\downarrow$4.65) &
  ($\downarrow$3.10) &
  ($\downarrow$2.97) &
  ($\downarrow$3.35) &
  ($\downarrow$3.54) &
  ($\downarrow$9.44) &
  ($\uparrow$2.01) &
  ($\downarrow$4.45) &
  ($\downarrow$3.86) &
  ($\downarrow$4.15) \\
\hline
\end{tabular}
}
\end{center}
\end{table}

\section{Generalization to Real world Use Cases}
\label{app-real-world-use-cases}
\subsection{Threat Reports to TTP Mapping}
\label{app-threat-to-ttp}
To assess the model’s ability to generalize to sources outside our training distribution, we built a benchmark using independently collected external threat-analysis reports, including technical write-ups, industry blogs, and vendor whitepapers. These documents were drawn from public sources not used in our training pipeline, allowing us to evaluate how well the model handles unseen threat reports.

Each report is paired with its corresponding attack technique through the existing mapping provided on the report’s associated campaign entry in public threat-intelligence repositories. Using this information, we formulate a multiple-choice classification task: the model receives the raw, unstructured report text and must select the correct technique from a set of candidate options. The model did not encounter this type of report-classification task during training, making it a novel reasoning setting in addition to the reports themselves being unseen.

We report CyberPal 2.0 performance on this benchmark in Table \ref{table:apt_cyberpal}.


\begin{table}[h]
    \caption{Threat reports to TTPs: CyberPal 2.0 models compared to their corresponding baselines.}
    \centering
    \begin{tabular}{lr @{\hskip 5pt} l} 
        \toprule
        \textbf{Model Name} & \multicolumn{2}{c}{\textbf{Score}} \\
        \midrule
        Qwen3-4B & 74.00 & \\
        \textbf{CyberPal-2.0-4B} & \textbf{82.75} & \gain{(+8.75)} \\
        \midrule
        Qwen 3 8b & 74.00 & \\
        \textbf{CyberPal-2.0-8B} & \textbf{78.25} & \gain{(+4.25)} \\
        \midrule
        Qwen 3 14b & 79.25 & \\
        \textbf{CyberPal-2.0-14B} & \textbf{85.25} & \gain{(+6.00)} \\
        \midrule
        gpt-oss-20B & 76.75 & \\
        \textbf{CyberPal-2.0-20B} & \textbf{79.50} & \gain{(+2.75)} \\
        \bottomrule
    \end{tabular}
    \label{table:apt_cyberpal}
    
    \vspace{0.1cm}
    \footnotesize
    \raggedright
\end{table}

    

\subsection{CyberSOCEval benchmark}
We further evaluate our models on \emph{CyberSOCEval} \cite{deason2025cybersoceval}, a recently released suite within CyberSecEval 4 that targets core SOC workflows. The benchmark comprises two multiple-choice tasks
\textbf{Malware Analysis} and \textbf{Threat Intelligence Reasoning} which scores models by exact-match accuracy (all and only the correct options) And Jaccard score (intersection of correct answers (the size of the intersection between the predicted and gold answer sets divided by the size of their union). 
\begin{itemize}
  \item \textbf{Malware Analysis} — Questions are grounded in real Sandbox detonation reports (e.g., process trees, extracted files, network activity). The task probes an LLM’s ability to interpret low-level telemetry and identify malicious behavior \cite{deason2025cybersoceval}.
  \item \textbf{Threat Intelligence Reasoning} — Questions are derived from full-page threat-intel reports (provided as page images), assessing an LLM’s capacity to extract actionable insights (e.g., adversary tactics, MITRE ATT\&CK mappings, targeted sectors) beyond surface-level comprehension \cite{deason2025cybersoceval}.
\end{itemize}

Since malware analysis tasks require extremely long context windows (up to 128k tokens for full prompts and approximately 32k for truncated ones), and our models were trained with a maximum sequence length of 8k tokens, we chose not to report results on this benchmark.
The full benchmark results are presented in Table~\ref{tables:soc_eval}.
To ensure consistent evaluation conditions, we re-ran all experiments using both LLaMA 4 and GPT-4o under identical settings, in particularly identical prompts which allows for a fair, apples-to-apples comparison across models.
Our results demonstrate that our model consistently outperforms the strongest baselines by a substantial margin, while remaining competitive with significantly larger open models, some up to ten times our model’s size narrowing the performance gap to roughly 4\% on average.

\begin{table}[h!]
\caption{CyberSocEval Threat Intelligence reasoning task}
\label{tables:soc_eval}
\begin{center}
\scalebox{0.9}{
\renewcommand{\arraystretch}{1.4} 
\begin{tabular}{l|cccccc}

        \textbf{Model}  &  Accuracy & Jaccard &  \\ 
        \hline
Qwen3-4B   & 5.95 & 10.12   \\
CyberPal-2.0-4B  & 19.39 & 51.85 \\
\hline
Qwen3-8B  & 42.86& 58.59  \\
CyberPal-2.0-8B   & 38.61 & 65.60 \\
\hline
Qwen3-14B   & 43.54 & 63.12 \\
CyberPal-2.0-14B & 45.07 & 67.78 \\
\hline

LLaMa-4-Maverick   & 54.25 & 72.57 \\
LLaMa-4-Scout  & 50.34 & 69.90  \\
GPT-4o & 53.57 & 73.19   \\
\hline
\end{tabular}
}
\end{center}
\end{table}

\subsection{CVE Reassessment benchmark}
\label{app-cve-reassess}
One real-world use case we are currently dealing with, both internally and with our clients, is how to reassess a CVE’s applicability and severity score for a specific package, deployment configuration, service, etc. 
 
Each CVE is associated with a Common Vulnerability Scoring System (CVSS) base score, which provides additional guidance about the vulnerability by scoring constant aspects such as: Attack Vector, Attack Complexity, User Interaction, Privileges Required, Scope, Confidentiality, Integrity, and Availability. 
 
A single CVE can affect many services and packages, and the CVSS score, derived from CVSS, often reflects a broad perspective and worst-case scenarios. In practice, CVSS base scores may vary for each vendor’s version, depending on the version they ship, how they ship it, the platform, and even how the software is compiled. This makes it difficult for third-party vulnerability databases (such as NVD), which can assign only a single CVSS base score per vulnerability, and also complicates comparison with vendors who score based on how the vulnerability manifests in their own products. 
 
Therefore, a key real-world use case is to reassess the CVSS score of a given CVE under specific constraints. For example, a NIST CVE may receive a High Impact score in general, but for a specific service or product that runs with low privileges, the effective Impact score may be lower. 
 
To study this, we constructed a benchmark of real CVEs (from 2025, to avoid data contamination) along with their original CVSS vectors. For each CVE, we paired specific products and services affected by that CVE and obtained new CVSS scores defined by security experts specifically for those products. The goal of this benchmark is to test how well LLMs can reassess the CVSS vector for concrete packages, services, and deployment configurations. 
 
Results are reported as MAD (Mean Absolute Deviation), the same metric used in CTI-Bench, normalized to the 0–1 range (1-[SCORE/10]). We also punish for bad responses which do not contain the final CVSS score by given the highest score of 10.
As can be seen from tables \ref{tab:cve_reassess_sec} and \ref{tab:cve_reassess_baseline}, our models show impressive improvements compared to both the baselines and other open-source cybersecurity LLMs. 

\begin{table}[h]
\caption{CVE Reassessment results of \textbf{open-source cybersecurity LLMs vs. CyberPal-2.0-8b.}}
\label{tab:cve_reassess_sec}
\begin{center}
\scalebox{0.9}{
\begin{tabular}{l|c}
    \toprule
    \textbf{Model Name} & \textbf{MAD (normalized)} \\
    \midrule
    Lily-Cyber-7B-v0.2 & 0.635  \\
    Llama-Primus-merged & 0.802 \\
    Llama-Primus-reasoning & 0.807 \\
    DeepHat-V1-7B & 0.823 \\
    Foundation-Sec-8B-Instruct & 0.698 \\
    \midrule
    \textbf{CyberPal-2.0-8B} & \textbf{0.834} \\
    \bottomrule
\end{tabular}
}
\end{center}
\end{table}

\begin{table}[h]
\caption{CVE Reassessment results of \textbf{CyberPal-2.0-8b vs. Baselines.}}
\label{tab:cve_reassess_baseline}
\begin{center}
\scalebox{0.9}{
\begin{tabular}{l|c}
    \toprule
    \textbf{Model Name} & \textbf{MAD (normalized)} \\
    \midrule
    Qwen3-4b & 0.783 \\
    CyberPal-2.0-4b & 0.830 \\
    \midrule
    Qwen3-8b & 0.825 \\
    CyberPal-2.0-8b & 0.834 \\
    \midrule
    Qwen3-14 & 0.740 \\
    CyberPal-2.0-14b & 0.833  \\
    \midrule
    gpt-oss-20b & 0.738 \\
    CyberPal-2.0-20b & 0.834 \\
    \bottomrule
\end{tabular}
}
\end{center}
\end{table}

\subsection{Secure Code Generation Benchmarks}
\label{app-sec-code-gen}
When developing a cybersecurity-oriented language model, an important risk emerges: models trained to reason about security may inadvertently generate insecure or vulnerability-prone code. This concern is especially relevant for LLMs designed to assist security practitioners, where users may rely on generated code for analysis, testing, or defensive automation. To ensure that our CyberPal 2.0 models do not introduce insecure coding patterns, we systematically evaluate them using established secure code generation benchmarks.

\textbf{CYBERSECEVAL} \citep{bhatt2023purple} introduces two complementary evaluation paradigms designed to measure how LLMs reproduce or generate insecure coding patterns in realistic development settings. Both benchmarks are built on a shared methodology: real-world insecure code is automatically identified in open-source repositories using the Insecure Code Detector (ICD), a static analysis framework containing 189 rules across 50 CWE categories, and these vulnerabilities are transformed into prompts that probe model behavior. By applying the same detection pipeline to model outputs, the benchmarks jointly characterize the security reliability of LLMs under different prompting modalities. 

\begin{itemize}
    \item \textbf{Insecure Code Generation — Autocomplete}
Evaluates whether a model continues code with insecure patterns when given preceding lines taken from insecure open-source snippets, reflecting risks in code-completion workflows.
    \item \textbf{Insecure Code Generation — Instruct}
Evaluates whether a model produces insecure code when responding to natural-language instructions generated from insecure code segments, reflecting risks in instruction-based coding workflows.
\end{itemize}

\begin{table}[h]
\caption{Secure Code Generation - Autocomplete, Instruct, and Average Pass Rates. 
Pass rate measures the percentage of test cases in which a model avoids reproducing insecure coding practices, as defined by the ICD.}
\centering
\begin{tabular}{lrrr}
\toprule
\textbf{Model Name} & \textbf{Autocomplete} & \textbf{Instruct} & \textbf{Average} \\
\midrule
Qwen-3-4B                      & 78.81 & 80.13 & 79.47 \\
CyberPal 2.0 Qwen-3-4B         & 76.10 & 67.16 & 71.63 \\
\midrule
Qwen-3-8B                      & 79.38 & 79.77 & 79.58 \\
CyberPal 2.0 Qwen-3-8B         & 72.81 & 69.07 & 70.94 \\
\midrule
Qwen-3-14B                     & 77.45 & 74.54 & 75.99 \\
CyberPal 2.0 Qwen-3-14B        & 73.33 & 66.33 & 69.83 \\
\midrule
gpt-oss-20b                    & 70.25 & 68.53 & 69.39 \\
CyberPal 2.0 gpt-oss-20b       & 67.90 & 63.95 & 65.93 \\
\bottomrule
\end{tabular}
\end{table}

Across all evaluated scales, CyberPal 2.0 demonstrates solid secure code generation performance, with results that remain close to those of the instruction-tuned comparison models - even though those models were further trained to provide safer responses, including in the context of secure code generation, while CyberPal 2.0 is trained directly from base models. On average, CyberPal 2.0 shows only a 6.52\% reduction relative to the aligned models, a modest change that aligns with expectations when adapting a model toward specialized cybersecurity reasoning. Despite this shift, CyberPal 2.0 retains most of the secure-coding characteristics of its reference models, indicating that the specialization process preserves core code-safety behavior while enabling substantially enhanced cybersecurity capabilities.

\section{Training time analysis}
To provide a clearer view of computational efficiency, Table~\ref{tab:training_time} reports a partial overview of the training durations for our models. 
All models were trained on NVIDIA A100 GPUs (80 GB) using a context length of 8192 tokens, a gradient accumulation step of 32, and an effective batch size of 3. 
We employed a boundary-preserving grouping and sequence-packing strategy to maximize hardware utilization and minimize idle time during training.

\begin{table}[h!]
\caption{Training time analysis}
\label{tab:training_time}
\begin{center}
\scalebox{0.9}{
\renewcommand{\arraystretch}{1.4} 
\begin{tabular}{l|c|c}

        Model size  &  GPU Count & Training hours  \\ 
        \hline
4B   & 40 & 1 Day, 6 hours and 40 minutes   \\
\hline
8B  & 40 & 2 Day, 19 hours and 18 minutes  \\

\hline
\end{tabular}
}
\end{center}
\end{table}

\section{Inference latency}
We benchmarked all model variants across quantization levels (FP16, 8-bit, 4-bit) and batch sizes (1, 4, 8) using a representative cyber security-style prompt (512 input tokens, 128 generated tokens) on a single NVIDIA H100-80GB GPU. 
For each configuration, we measured end-to-end generation latency, approximate time-to-first-token (TTFT), throughput (tokens per second), and peak GPU memory footprint (including weights and KV-cache).
Results are reported in seconds to highlight practical latency ranges. 
The measurements show that FP16 inference, as expected has the highest token per sec rate, while 8-bit quantization reduces memory usage by nearly 40\% with moderate throughput trade-offs. Batch scaling demonstrates nearly linear throughput gains until GPU saturation, indicating stable memory behavior across configurations. These findings confirm that quantized models can meet real-time operational requirements (e.g.,  less then 100 ms per token generation) while significantly reducing hardware cost.

Table~\ref{tab:perf_metrics} summarizes the observed performance. 
Our 8-bit variants consistently exhibited higher end-to-end latency compared to both 4-bit and FP16 configurations.
While their overall throughput (tokens per second) remained competitive, the TTFT was substantially longer—often by an order of magnitude.
Although we cannot definitively isolate the cause, prior work suggests that some hardware backends perform runtime de-quantization of 8-bit weights \cite{zhang2024mixpequantizationhardwarecodesign}, introducing additional computational overhead. Moreover, a recent large-scale empirical study reported that “quantization does not always reduce latency in online serving.” \cite{shi2025systematic}.
Together, these observations explain why our 8-bit inference runs showed higher latency than FP16 despite achieving a smaller memory footprint.
\begin{table}[h!]
\centering
\caption{Inference latency}
\label{tab:perf_metrics}
\begin{tabular}{llrrrrrrrr}
\toprule
   \multicolumn{1}{l}{\begin{tabular}[c]{@{}l@{}}Model\\ Label\end{tabular}} &
   \multicolumn{1}{l}{\begin{tabular}[c]{@{}l@{}}Bit\\ Mode\end{tabular}} &
   \multicolumn{1}{l}{\begin{tabular}[c]{@{}l@{}}Batch\\ Size\end{tabular}} &
   \multicolumn{1}{l}{\begin{tabular}[c]{@{}l@{}}Prompt\\ Length\end{tabular}} &
   \multicolumn{1}{l}{\begin{tabular}[c]{@{}l@{}}Generation\\ Length\end{tabular}} &
   \multicolumn{1}{l}{\begin{tabular}[c]{@{}l@{}}Latency\\ (Sec)\end{tabular}} &
   \multicolumn{1}{l}{\begin{tabular}[c]{@{}l@{}}TTFT \\ (Sec)\end{tabular}} &
   \multicolumn{1}{l}{\begin{tabular}[c]{@{}l@{}}Tokens \\ Per Sec \end{tabular}} &
   \multicolumn{1}{l}{\begin{tabular}[c]{@{}l@{}}Memory  \\ peak (GB) \end{tabular}}&
    \\
  
\midrule
 CyberPal-4B &     4bit &           1 &         512 &      128 &      5.34 &          0.042 &                 23.95 &        3.27  \\
 CyberPal-4B &     4bit &           4 &         512 &      128 &      7.83 &          0.060 &                 65.39 &        3.37 \\
 CyberPal-4B &     4bit &           8 &         512 &      128 &      7.64&          0.06 &                134.07 &        3.46  \\
 CyberPal-4B &     8bit &           1 &         512 &      128 &     22.45 &          0.175 &                  5.70 &        4.83 &   \\
 CyberPal-4B &     8bit &           4 &         512 &      128 &     22.83 &          0.178 &                 22.43 &        4.89 \\
 CyberPal-4B &     8bit &           8 &         512 &      128 &     18.60 &          0.145 &                 55.05&        4.99  \\
 CyberPal-4B &     fp16 & 1 & 512 &      128 &      4.09 &          0.032 &                 31.33 &        8.440  \\
 CyberPal-4B &     fp16 &           4 &         512 &      128 &      9.68 &          0.076 &                 52.91 &        8.52 \\
 CyberPal-4B &     fp16 &           8 &         512 &      128 &      4.86 &          0.038 &                210.52 &        8.62  \\
 \hline 

 CyberPal-8B &     4bit &           1 &         512 &      128 &      5.30 &          0.041 &                 24.15 &        6.53  \\
 CyberPal-8B &     4bit &           4 &         512 &      128 &      7.56 &          0.059 &                 67.72 &        6.62  \\
 CyberPal-8B &     4bit &           8 &         512 &      128 &      7.55 &          0.059 &                135.70 &        6.72 \\
 CyberPal-8B &     8bit &           1 &         512 &       71 &      8.23 &          0.116 &                  8.63 &        9.55  \\
 CyberPal-8B &     8bit &           4 &         512 &      128 &     16.42 &          0.128 &                 31.18 &        9.58 \\
 CyberPal-8B &     8bit &           8 &         512 &      128 &     16.97 &          0.133 &                 60.33 &        9.66 \\
 CyberPal-8B &     fp16 &           1 &         512 &       59 &      1.87 &          0.032 &                 31.53 &       18.83\\
 CyberPal-8B &     fp16 &           4 &         512 &       59 &      1.90 &          0.032 &                124.15 &       18.88\\
 CyberPal-8B &     fp16 &           8 &         512 &       59 &      1.89 &          0.032 &                249.78 &       18.95  \\ 
 \hline
 CyberPal-14B &     4bit &           1 &         512 &       21 &      0.99 &          0.047 &                 21.17 &       14.69  \\
CyberPal-14B &     4bit &           4 &         512 &       21 &      1.99 &          0.095 &                 42.23 &       14.72  \\
CyberPal-14B &     4bit &           8 &         512 &       21 &      1.87 &          0.089 &                 89.90 &       14.76\\  

CyberPal-14B &     8bit &           1 &         512 &      128 &     19.15 &          0.150 &                  6.68 &       20.48 \\
CyberPal-14B &     8bit &           4 &         512 &      128 &     19.34 &          0.151 &                 26.48 &       20.51 \\
CyberPal-14B &     8bit &           8 &         512 &      128 &     18.62 &          0.145 &                 55.01 &       20.55 \\
CyberPal-14B &     fp16 &           1 &         512 &      128 &      5.790  &         0.045 &                 22.11 &       33.50 \\
CyberPal-14B &     fp16 &           4 &         512 &      128 &      7.36 &          0.057 &                 69.59 &       33.58 \\
CyberPal-14B &     fp16 &           8 &         512 &      128 &      5.26 &          0.041 &                194.78 &       33.69 \\
\hline
CyberPal-20B &     4bit &           1 &         512 &      128 &      5.06 &          0.04 &                 25.31 &       40.98  \\
CyberPal-20B &     4bit &           4 &         512 &      128 &      5.62 &          0.044 &                 91.14 &       41.15  \\
CyberPal-20B &     4bit &           8 &         512 &      128 &      5.55 &          0.043 &                184.39 &       41.37   \\
CyberPal-20B &     8bit &           1 &         512 &      128 &     10.96 &          0.086 &                 11.68 &       41.26 \\
CyberPal-20B &     8bit &           4 &         512 &      128 &     11.90 &          0.093 &                 43.03 &       41.43  \\
CyberPal-20B &     8bit &           8 &         512 &      128 &     13.34 &          0.104 &                 76.79 &       41.65 \\
CyberPal-20B &     fp16 &           1 &         512 &      128 &      4.54 &          0.035 &                 28.18 &       49.95  \\
CyberPal-20B &     fp16 &           4 &         512 &      128 &      5.56 &          0.043 &                 92.01 &       50.11  \\
CyberPal-20B &     fp16 &           8 &         512 &      128 &      5.890&          0.046 &                173.9 &       50.34 \\

\bottomrule
\end{tabular}
\end{table}

\section{Use of Large Language Models (LLMs)}
When writing the paper, we used LLMs to help us find grammar errors and polish sentences that needed further clarifications. No further usage was done using LLMs while writing the paper. 

\section{Qualitative results}

\begin{figure*}[h!]
  \centering
  \includegraphics[width=\textwidth]{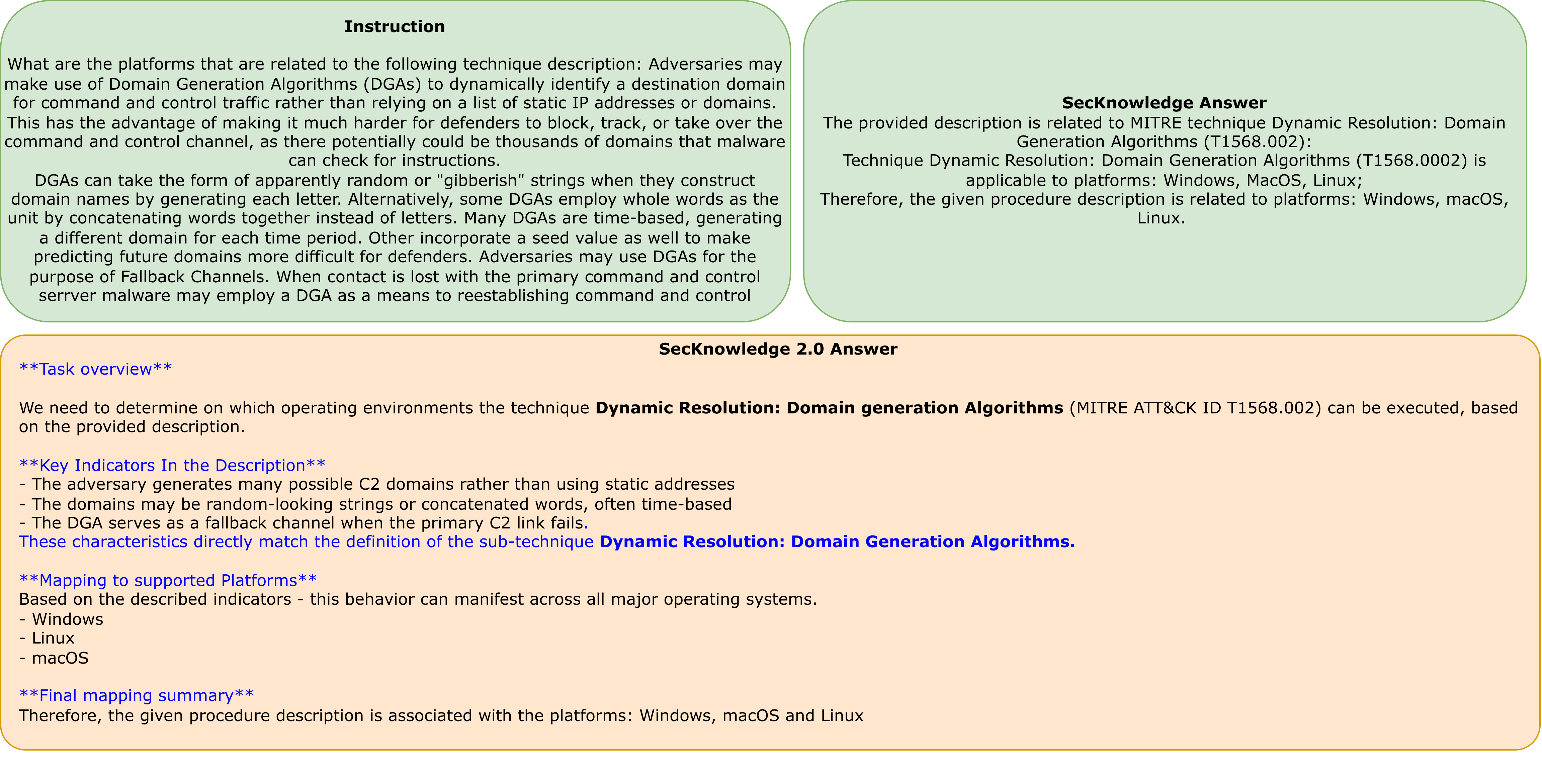}
  \caption{Instruction (green) Q\&A example from SecKnowledge (green), and our improved answer (orange).}
  \label{figures:realign-example-2}
\end{figure*}

\begin{figure*}[h!]
  \centering
  \includegraphics[width=\textwidth]{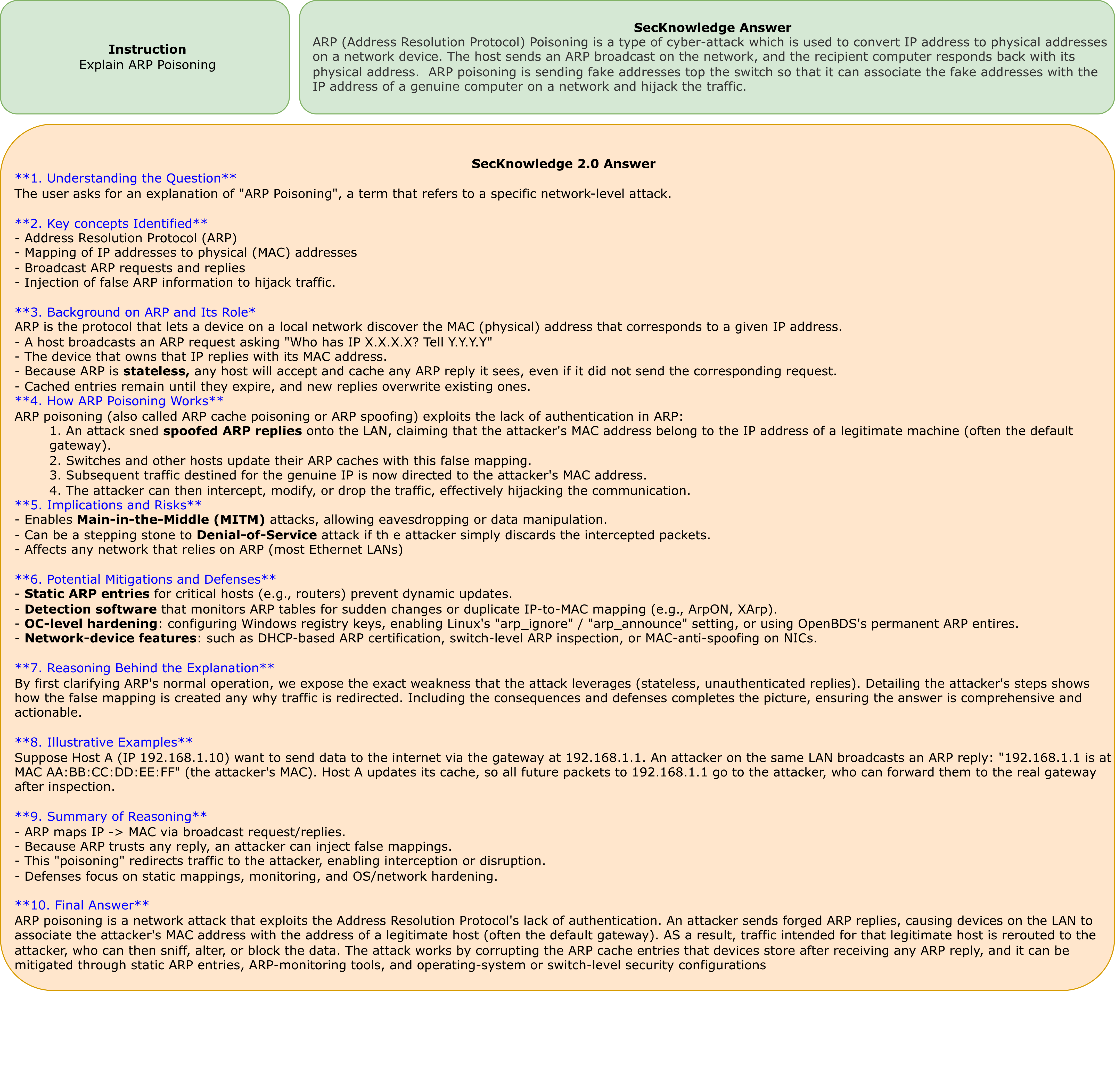}
  \caption{Instruction (green) Q\&A example from SecKnowledge (green), and our improved answer (orange).}
  \label{figures:realign-example-1}
\end{figure*}

We include additional examples of the training examples before and after pipeline in Figure~\ref{figures:realign-example-2} and Figure~\ref{figures:realign-example-1}

\end{document}